\documentclass[11pt]{article}

\usepackage[preprint]{acl}

\usepackage{times}
\usepackage{latexsym}

\usepackage[T1]{fontenc}
\usepackage[utf8]{inputenc}

\usepackage{microtype}

\usepackage{inconsolata}

\usepackage{graphicx}
\usepackage{wrapfig}
\usepackage[ruled,vlined]{algorithm2e}
\usepackage{float}
\usepackage[table]{xcolor}

\definecolor{headergray}{gray}{0.85}
\usepackage[most]{tcolorbox}
 \usepackage{caption}
\usepackage{amssymb}
\usepackage{tabularx}

\usepackage{hyperref}
\usepackage{url}
\usepackage{booktabs}
\usepackage{adjustbox}

\title{A Comparative Analysis of Sparse Autoencoder and Activation Difference in Language Model Steering}

\author{Jiaqing Xie \\
  Department of Computer Science\\
  ETH Zurich \\
  \texttt{jiaxie@ethz.ch}}

\begin{document}
\maketitle
\begin{abstract}
Sparse autoencoders (SAEs) have recently emerged as a powerful tool for language model steering. Prior work has explored top-k SAE latents for steering, but we observe that many dimensions among the top-k latents capture non-semantic features such as punctuation rather than semantic attributes like instructions. To address this, we propose focusing on a single, most relevant SAE latent (top-1), eliminating redundant features. We further identify a limitation in constant SAE steering, which often produces degenerate outputs such as repetitive single words. To mitigate this, we introduce a token-wise decaying steering strategy, enabling more faithful comparisons with mean activation difference baselines. Empirically, we show that steering an SAE latent associated with reasoning reliably elicits step-by-step mathematical reasoning and enhances inference quality, functionally resembling the effect of appending a guiding token. Our results demonstrate that SAEs outperform mean activation difference methods on mathematical reasoning benchmarks and match their performance on IF-Eval.
\end{abstract}

\section{Introduction}

\begin{figure*}[!htb]
    \centering
    \includegraphics[width=\linewidth]{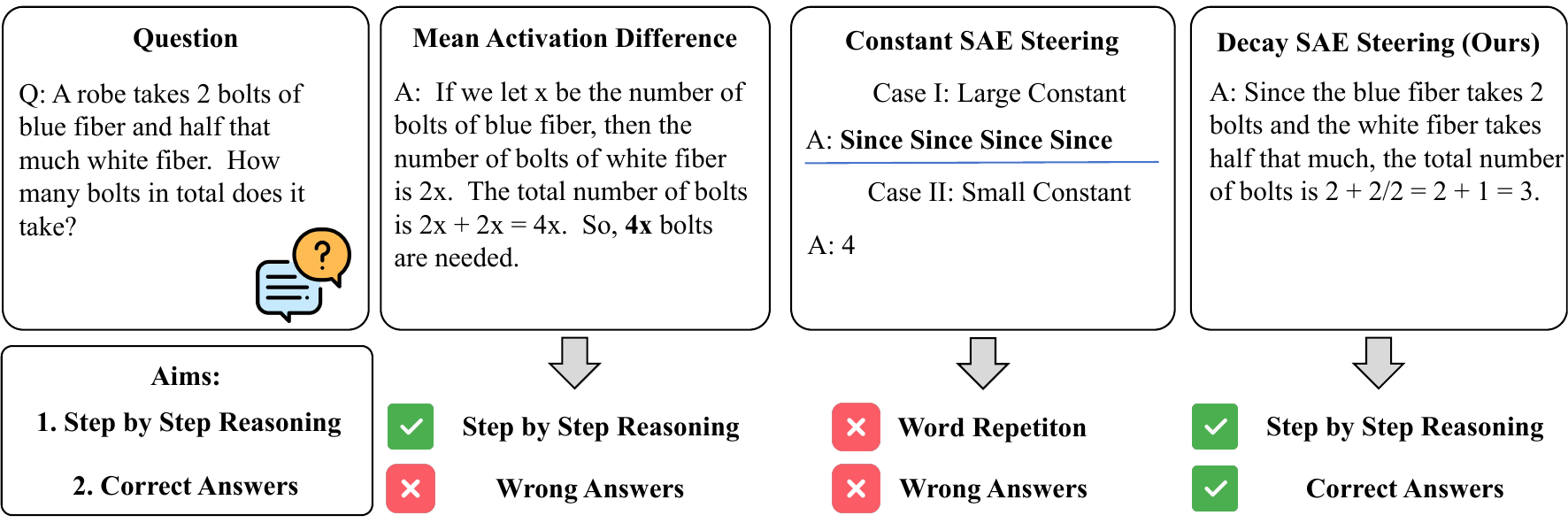}
    \caption{Mean Activation Difference (MeanActDiff) produces incorrect answers in math reasoning cases, while constant SAE steering often leads to word repetitions or wrong outputs. Our proposed decaying strategy applies the SAE steering vector progressively, yielding both step-by-step reasoning and correct answers, and enabling a fair comparison with MeanActDiff.}
    \label{fig:pipeline}
\end{figure*}

Steering language models by directly manipulating internal activations is an increasingly popular method for controlling model behavior without fine-tuning or prompt engineering~\citep{stolfo2024improving, postmus2024steering}
. Methods such as mean activation difference (\textbf{MeanActDiff})~\citep{stolfo2024improving} and sparse autoencoder (\textbf{SAE})~\citep{makelov2024sparse, mayne2024sparseautoencodersuseddecompose, obrien2024steeringlanguagemodelrefusal}-based steering have shown initial success in tasks including sentiment control~\citep{tigges2024language} and style transfer~\citep{turner2023activation, liu2024incontext, scalena-etal-2024-multi, vonrütte2024languagemodelsguidelatent}. Recently, some works have extended the scenarios to complex math reasoning and instruction-following tasks ~\citep{galichin2025have, wu2025axbench}, demonstrating the robustness and versatility of SAE-based steering.

However, MeanActDiff lacks interpretability, offering little insight into the manipulated semantic features~\citep{liu2024incontext}. SAE-based steering provides a more interpretable alternative by exposing sparse, human-understandable activation features~\citep{bloom2024saetrainingcodebase}. Existing SAE steering methods often introduce noise and redundancy by selecting top-activated features without verifying their relevance~\citep{shu2025survey, rajamanoharan2024improving}. In addition, most existing methods apply a constant steering strength throughout generation, which risks destabilizing generation or causing unnatural repetition, particularly in multi-step reasoning tasks, as shown in Figure~\ref{fig:pipeline}. These limitations have prevented steering methods from being broadly adopted in more challenging scenarios, such as chain-of-thought math reasoning~\citep{wei2022chain} and fine-grained instruction~\citep{zhou2023instruction} adherence, where precise and stable control is essential.

To address these challenges, we first propose an SAE feature selection mechanism based on subtracting top-activated feature sets, which isolates the dimensions most responsible for behavioral shifts while filtering out irrelevant activations. Second, we design a token-wise decaying steering strategy that dynamically reduces intervention strength during generation, improving stability and mitigating oversteering. We evaluate these methods on Gemma-2-2b and Gemma-2-9b for chain-of-thought math reasoning (GSM8K~\citep{cobbe2021training}, SVAMP~\citep{patel-etal-2021-nlp}, MAWPS~\citep{koncel-kedziorski-etal-2016-mawps}, ASDIV~\citep{miao-etal-2020-diverse}) and on Gemma-2-9b-it for instruction following (IFEval~\citep{zhou2023instruction}, covering JSON formatting, case sensitivity, multilingual responses, and lexical constraints). Our results show that decaying SAE steering improves stability and achieves finer control compared to both MeanActDiff and constant SAE steering. In math reasoning, SAE steering reliably elicits step-by-step solutions without requiring few-shot prompts, outperforming MeanActDiff. In instruction following, our feature selection method more clearly identifies redundant latents, while in math reasoning further refinement may be needed. Finally, we observe that SAE steering behaves similarly to appending an additional guiding token to the prompt.

\section{Related Works}
\paragraph{Language Model Steering via Activation Editing
}
Recent research into mechanistic interpretability explores direct control over language model outputs by intervening in their internal workings~\citep{conmy2023towards, rai2024practical}. This is achieved by adding or subtracting steering vectors from hidden activations at specific layers to influence the style, tone, or content of the generated output~\citep{subramani-etal-2022-extracting, chalnev2024improving}.
A primary method for identifying these directions is the Mean Activation Difference (MeanActDiff)~\citep{stolfo2024improving}, which computes the average difference in activations between input pairs that elicit contrasting behaviors (e.g., generic vs. instruction-following responses)~\citep{liu2024context}. Several variants such as MeanActDiff-PCA and Linear Artificial Tomography (LAT) have been developed ~\citep{zou2025representationengineeringtopdownapproach, wu2025axbench}.

\paragraph{Sparse Autoencoders for Intepretable Steering}
To overcome the lack of interpretability in MeanActDiff's global steering vectors, Sparse Autoencoders (SAEs) offer a more expressive alternative by learning a compressed and disentangled representation of a model's hidden activations~\citep{lieberum2024gemma, gao2024scaling}. SAE training has followed diverse directions. Some works like Gated, Top-K and Switch SAE are trying to improve the SAE architecture~\citep{gao2024scaling,rajamanoharan2024improving,mudide2024efficient}, while others like LayerGroup, and Formal Language SAE~\citep{ghilardi2024efficient,menon2024analyzing} are trying to improve the training strategies.
These pre-trained SAEs~\citep{lieberum2024gemma,he2024llama,chaudhary2024evaluatingopensourcesparseautoencoders} have enabled applications such as steering semantic instructions~\citep{wu2025interpreting}, biological features~\citep{farrell2024applying}, reveal the hallucinations in LLMs~\citep{ferrando2024know}, and study cross-lingual toxicity behaviours~\citep{gallifant2025sparse}.
\section{Methods}
\begin{figure*}[!htb]
    \centering
    \includegraphics[width=0.9\linewidth]{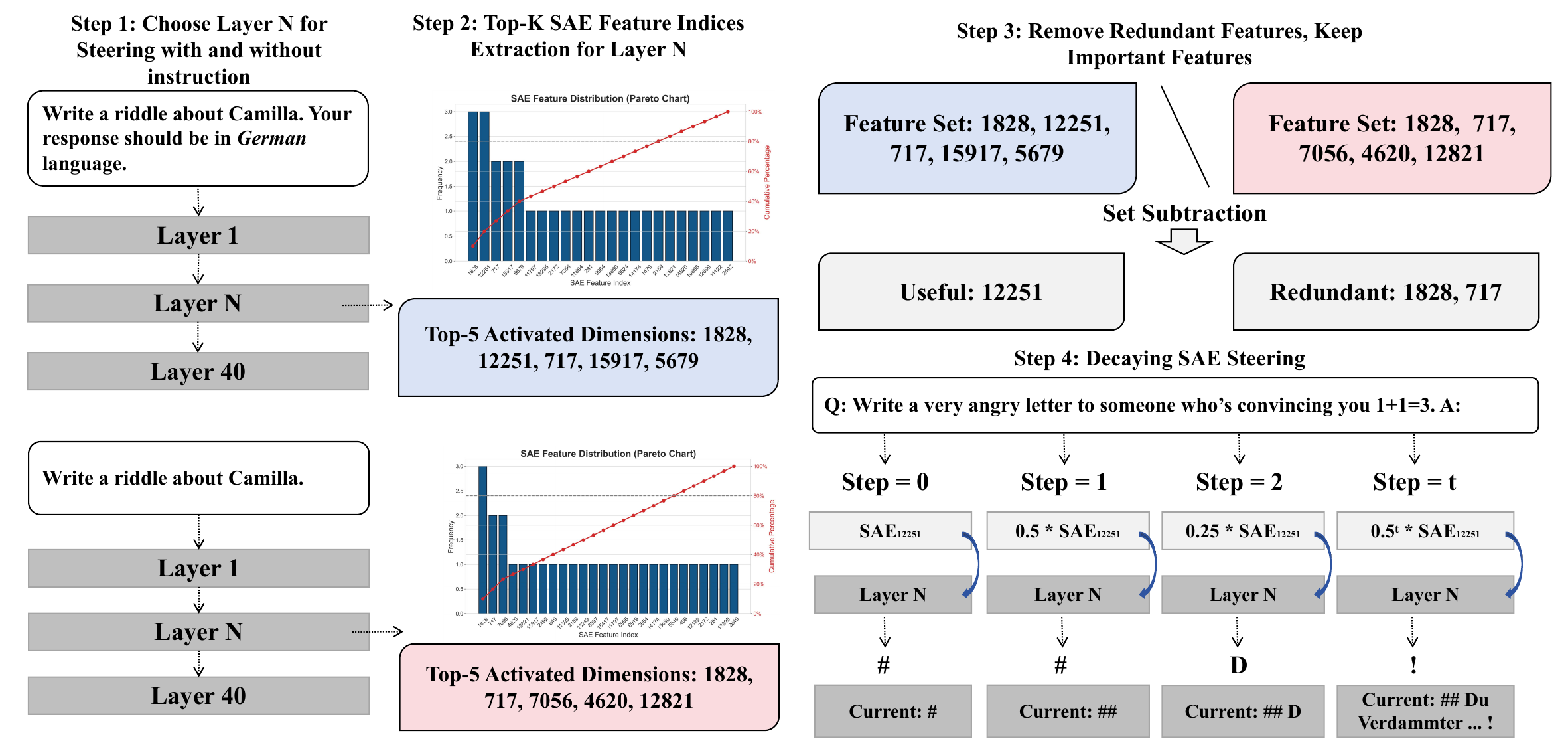}
    \caption{Pipeline of SAE feature extraction and token-wise decaying steering applied to question answering in \textbf{German}.}
    \label{fig:SAE_feature_extraction_and_steering}
\end{figure*}

% \subsection{Problem Formulation}
% While AxBench~\citep{wu2025axbench} provides a foundational comparison between MeanActDiff and SAE steering methods, its evaluation framework exposes two key limitations that this work aims to address.
% First, the evaluation of SAEs does not account for the influence of irrelevant or non-causal latent features. The inclusion of these features during steering may fail to improve performance, keeping answers unchanged. This necessitates a method for isolating the subset of SAE features that are causally linked to the desired behavior.
% Second, a limitation arises from the prevalent use of \textit{constant steering} in prior work~\citep{chalnev2024improving,obrien2024steeringlanguagemodelrefusal,yang2025lf}. We observe that applying a static steering vector at each generation step can induce undesirable artifacts, most notably token-level repetition and stylistic rigidity. This suggests that the effectiveness of steering may depend on a more dynamic application strategy.
% These two observations highlight the need for a more rigorous and equitable comparison between activation difference and SAE-based steering methodologies. To address these gaps, this section first introduces the necessary notation and formalizes the baseline methods for MeanActDiff and SAE steering. Subsequently, we detail our proposed contributions: a method for extracting task-relevant SAE features (Section 3.3) and an adaptive steering strategy designed to mitigate generative artifacts (Section 3.4).

\subsection{Preliminaries and Notation}
Let $\mathcal{T} = \{\mathcal{T}_i\}_{i=0}^{K-1}$ denote a test set comprising $K$ instances. Correspondingly, let $\mathcal{T}^* = \{\mathcal{T}^*_i\}_{i=0}^{K-1}$ be an augmented version of this set, where each instance is modified to elicit a desired behavior. In the context of mathematical reasoning, this augmentation is primarily achieved through in-context learning (ICL), where few-shot chain-of-thought examples are prepended to the input. For instruction following cases, we appended the instructions to the input as well. Let $\mathcal{F}$ represent a pretrained large language model. We define $\mathcal{A}_l(\cdot)$ as an operator that extracts the hidden activation vector (e.g., from an MLP or attention layer) from layer $l$ of $\mathcal{F}$ for a given input. Furthermore, let $\{\mathcal{S}_l\}$ be a set of pretrained Sparse Autoencoders, where each $\mathcal{S}_l$ is trained on the activations from layer $l$ of $\mathcal{F}$. Each SAE contains a dictionary of interpretable features, $\{\mathbf{d}_{l,j}\}_{j=1}^{N_l}$, where $\mathbf{d}_{l,j}$ is the $j$-th feature vector (dictionary atom) for layer $l$ and $N_l$ is the total number of learned features. With this formalism, we can define the two primary steering methods.

\paragraph{Mean Activation Difference (MeanActDiff) Vector}
The MeanActDiff method generates a global steering vector by averaging the activation differences between paired examples from $\mathcal{T}^*$ and $\mathcal{T}$. For a given layer $l$, the MeanActDiff steering vector, $\mathbf{v}_l$, is formally defined as:
\begin{equation}
    \mathbf{v}_l = \frac{1}{K}\sum_{i=0}^{K-1} \left( \mathcal{A}_l(\mathcal{F}(\mathcal{T}^*_i)) - \mathcal{A}_l(\mathcal{F}(\mathcal{T}_i)) \right)
\end{equation}
This vector represents the average direction in activation space that corresponds to the behavior induced by the instructions or examples in $\mathcal{T}^*$.

\paragraph{Sparse Autoencoder (SAE) Features}
In contrast to deriving a vector from behavioral differences, a Sparse Autoencoder (SAE) discovers an overcomplete basis of interpretable features directly from the distribution of a model's activations. For a given layer $l$, an SAE is an encoder-decoder network trained to reconstruct the activation vector $\mathbf{h}_l = \mathcal{A}_l(\mathcal{F}(\mathcal{T}^*_i))$. The encoder maps the activation $\mathbf{h}_l$ to a high-dimensional, sparse latent vector $\mathbf{z}_l \in \mathbb{R}^{d_{sae}}$ (where $d_{sae} \gg d_{model}$):
\begin{equation}
    \mathbf{z}_l = \text{ReLU}(\mathbf{W}_e \mathbf{h}_l + \mathbf{b}_e)
\end{equation}
where $\mathbf{W}_e \in \mathbb{R}^{d_{sae} \times d_{model}}$ is the encoder weight matrix and $\mathbf{b}_e$ is the encoder bias. 
The decoder then reconstructs the original activation from the sparse representation $\mathbf{z}_l$:
\begin{equation}
    \hat{\mathbf{h}}_l = \mathbf{W}_d \mathbf{z}_l + \mathbf{b}_d
\end{equation}
where $\mathbf{W}_d \in \mathbb{R}^{d_{model} \times d_{sae}}$ is the decoder weight matrix and $\mathbf{b}_d$ is the decoder bias. Crucially for steering, the columns of the trained decoder matrix $\mathbf{W}_d$ form an overcomplete dictionary of feature vectors, $\{\mathbf{d}_{l,j}\}_{j=1}^{d_{sae}}$. Each vector $\mathbf{d}_{l,j}$ represents a distinct, approximately monosemantic feature learned by the SAE. This provides a rich dictionary of potential steering vectors, from which one or more can be selected to guide generation.

\subsection{Most Relevant SAE Feature Extraction}

\begin{algorithm}[!htb]
\caption{Differentially Activated Feature Extraction}
\label{alg:diff_activated_features}
\KwIn{LLM $\mathcal{F}$, SAE Encoder $\mathcal{S}_{l,enc}$, layer index $l$, datasets $\mathcal{T}, \mathcal{T}^*$, integer $K$}
\KwOut{Set of differentially activated SAE feature indices}

\SetKwBlock{Begin}{Initialize:}{}
\Begin{
    $\mathbf{S}, \mathbf{S}^* \gets \{\}, \{\}$ 
    % \text{Frequency counters for top-K indices}
}

\For{$i \gets 0$ \KwTo $K-1$}{
    % 1. Get the full sequence of activations for each instance
    $\mathbf{H}_{l}, \mathbf{H}^*_{l} \gets \mathcal{A}_l(\mathcal{F}(\mathcal{T}_i)), \mathcal{A}_l(\mathcal{F}(\mathcal{T}^*_i))$ 

    % 2. Encode the entire sequence of activations
    $\mathbf{Z}_{l}, \mathbf{Z}^*_{l} \gets \mathcal{S}_{l,enc}(\mathbf{H}_{l}), \mathcal{S}_{l,enc}(\mathbf{H}^*_{l})$ 
    % \text{Matrix of shape (seq\_len, d\_sae)} \;
    
    % 3. Select the sparse code vector for the final token
    $\mathbf{z}_{l, \text{last}}, \mathbf{z}^*_{l, \text{last}} \gets \mathbf{Z}_{l}[-1], \gets \mathbf{Z}^*_{l}[-1]$\;

    % 4. Find top-K activated feature indices from the final token's sparse code
    $\text{indices}, \text{indices}^* \gets \text{TopK}(\mathbf{z}_{l, \text{last}}, K), \text{TopK}(\mathbf{z}^*_{l, \text{last}}, K)$\;
    
    % Update frequency counters
    \ForEach{idx, $\text{idx}^*$ $\in \text{paired(indices, }\text{indices}^*\text{)}   $}{
        $\mathbf{S}[\text{idx}] \gets \mathbf{S}.get(\text{idx}, 0) + 1$\;
        $\mathbf{S}^*[\text{idx}^*] \gets \mathbf{S}^*.get(\text{idx}, 0) + 1$\;
    }
}
\Return $\text{RankedKeys}(\mathbf{S}^*) \setminus \text{RankedKeys}(\mathbf{S})$
\end{algorithm}

To isolate the SAE features causally linked to the behavior elicited by the augmented dataset $\mathcal{T}^*$, we propose a method to identify features that are differentially activated when the model processes $\mathcal{T}^*$ compared to the standard dataset $\mathcal{T}$. The objective is to find features that activate consistently for the augmented (e.g., chain-of-thought) inputs while remaining inactive for the standard inputs, thereby pinpointing the representations of the desired behavior. Our procedure, formalized in Algorithm~\ref{alg:diff_activated_features}, focuses on the model's state at the most critical juncture: the final token position of the input prompt. This pre-generation state is highly influential for the model's response.

\paragraph{Algorithm Description}
Algorithm~\ref{alg:diff_activated_features}  iterates through the paired instances from the paired datasets $\mathcal{T}$ and $\mathcal{T}^*$. For each pair $(\mathcal{T}_i, \mathcal{T}^*_i)$, we perform the following steps:
\begin{enumerate}
    \item We run the model $\mathcal{F}$ on the entire input sequence and use the operator $\mathcal{A}_l(\cdot)$ to extract the full sequence of hidden activations at a target layer $l$. This results in two activation matrices, $\mathbf{H}_l$ and $\mathbf{H}^*_l$, where each matrix has dimensions (sequence length $\times$ $d_{model}$).
    \item Each activation matrix is passed through the encoder of the corresponding SAE, $\mathcal{S}_l$, to produce a sequence of sparse latent representations, $\mathbf{Z}_l$ and $\mathbf{Z}^*_l$.
    \item From these latent representation matrices, we select only the vector corresponding to the \textbf{final token}. Let these vectors be $\mathbf{z}_{l, \text{last}}$ and $\mathbf{z}^*_{l, \text{last}}$.
    \item We select indices of the top-$K$ highest-activated SAE features within these final-token vectors.
\end{enumerate}
Two frequency counters, $\mathbf{S}$ and $\mathbf{S}^*$, track how often each feature index appears among the top-$K$ features across $\mathcal{T}$ and $\mathcal{T}^*$, respectively. The final set of candidate features consists of those with substantially higher frequency in $\mathcal{T}^*$ than in $\mathcal{T}$, thereby highlighting the features most associated with the augmented behavior.

\subsection{Steering Strategy}

\paragraph{Constant Steering}
The simplest strategy is constant steering. This strategy involves consistently adding a fixed steering vector $\mathbf{v}_{steer}$ to the hidden activation $\mathbf{h}_l$ at a target layer $l$ during every step of the auto-regressive process. The steered activation, $\mathbf{h}'_l$, is computed as:
\begin{equation}
    \mathbf{h}'_l = \mathbf{h}_l + \alpha \cdot \mathbf{v}_{steer}
\end{equation}
Here, $\mathbf{v}_{steer}$ can be a MeanActDiff vector $\mathbf{v}_l$ or a selected SAE feature $\mathbf{d}_{l,j}$, and $\alpha$ is a constant scaling factor that controls the steering strength. To ensure effective influence on the output, this modification is typically applied by splitting the model's forward pass, often implemented efficiently using hooks or activation patching. While effective for persistent behaviors like maintaining a specific sentiment, this constant intervention can lead to artifacts such as token repetition ("Since Since Since...") or overly rigid outputs in more complex tasks, as shown in Figure~\ref{fig:pipeline}.

\paragraph{Adaptive Steering: A Decaying Approach}
To mitigate the issues of constant steering, particularly in multi-step reasoning tasks, we propose a more adaptive \textbf{decaying steering} strategy.  The core idea is to make the steering influence strong at the beginning of generation and then diminish it over time. This allows the intervention to guide the model's initial path (e.g., to start a chain-of-thought process) before gracefully transitioning control back to the model's own learned priors for subsequent steps. This is achieved by modifying the update rule to use a time-dependent scaling factor, $\alpha_t$:
\begin{equation}
    \mathbf{h}'_l = \mathbf{h}_l + \alpha_t \cdot \mathbf{v}_{steer}, \quad \alpha_t = \alpha \cdot \left(\frac{1}{1 + \omega \cdot t}\right)^k
\end{equation}
In this formulation, $\alpha$ is the initial base weight, $t$ is the current generation step (starting from $t=0$), $\omega$ is a hyperparameter controlling the rate of decay, and $k$ is a decay exponent that adjusts the steepness of the falloff. Constant steering applies a fixed $\alpha$, whereas we use a time-dependent $\alpha_t$. To maintain representational stability and prevent distortions in downstream layers, the steered activation $\mathbf{h}'_l$ is typically normalized to match the L2-norm of the original activation $\mathbf{h}_l$ before being passed onward, which is $\mathbf{h}'_l * \|\mathbf{h}_l\|_2/\|\mathbf{h}'_l\|_2$. This dynamic approach alleviates the repetitiveness of constant steering and proves more effective for tasks requiring generative flexibility~\citep{liu2024incontext}.

\section{Experiments}
All experiments are conducted on two RTX 3090 GPUs. To optimize memory usage while maintaining numerical precision, we utilize the bfloat16 data type during both inference and activation recording phases for Gemma-2-9B and Gemma-2-9B-it models. For the Gemma-2-2B model, we retain full 32-bit floating-point precision. We fix the temperature to 0 and set top-p to 1, and conduct the experiment with a single run.
\paragraph{Datasets}
We split the experiment environment into two settings: 1) chain-of-thought math reasoning and 2) instruction following. In the first setting, we test our methods on GSM8K~\citep{cobbe2021training}, SVAMP~\citep{patel-etal-2021-nlp}, MAWPS~\citep{koncel-kedziorski-etal-2016-mawps}, and ASDIV~\citep{miao-etal-2020-diverse} benchmarks. In the second setting, we test our methods on the IFEval~\citep{zhou2023instruction} benchmark subsets including JSON formatting, uppercase and lowercase generations, different language responses and word inclusions.

\paragraph{Train-Test Split}
For language-specific cases, the available training examples per language are limited (fewer than 30). Consequently, we allocate 30\% of the data for SAE steering and MeanActDiff steering, 20\% for tuning hyperparameters related to steering strategies, and the remaining 50\% for inference. For all other cases in the IFEval dataset and math reasoning datasets, we split the data into 50\% for training, 20\% for testing, and 30\% for validation. We perform token-wise generation control with \textit{max\_new\_token} set to 1. Additionally, we set \textit{top\_p} to 1 and use a low \textit{temperature} of 0.05 to encourage more deterministic outputs.

\paragraph{Inference}
For standard input, MeanActAdd steering, and SAE steering, we use zero-shot inference. For chain-of-thought (CoT) inference in math reasoning tasks, we use few-shot settings with 8 CoT examples. In Instruction following tasks, we append the instruction to the end of the input.

% \begin{wrapfigure}{r}{0.45\textwidth}
%     \vspace{-20pt} 
%     \centering
%     \caption{Models and their corresponding layer index for steering.}
%     \label{tab:model_layer_index}
%     \vspace{5pt}
%     \begin{tabular}{lcc}
%         \toprule
%         \textbf{Model} & \textbf{Layer} & \textbf{Total Layers} \\ 
%         \midrule
%         Gemma-2-2B    & 20 & 26 \\ 
%         Gemma-2-9B    & 31 & 42 \\ 
%         Gemma-2-9B-it & 31 & 42 \\ 
%         \bottomrule
%     \end{tabular}
% \end{wrapfigure}

\paragraph{Layer Choices for SAE Importance Analysis}
We select a specific transformer layer from each model for SAE extraction and steering experiments.  The chosen layers correspond to around four-fifth of the total number of layers in the model, where semantic abstraction tends to stabilize and steerable latents are more linearly separable, which are 20 for Gemma-2-2b, and 31 for Gemma-2-9b and its instruction-tuned version.

\subsection{Hyper-parameters of Steering}
%Tables~\ref{tab:gemma_hyperparam_cases} and~\ref{tab:gemma_hyperparam_cases_IFEval} list the hyperparameters used across different steering settings. Each case (e.g., format control, length steering) includes both MeanActDiff and SAE-based methods applied to the same base model. Hyper-parameters include the scaling factor $\omega$, decaying factor $K$, and steering weight, all manually tuned to balance output controllability and semantic coherence.  

For the Instruction following dataset IFEval, we use a non-decaying factor ($k = 0$ which follows the setting of activation steering on IFEval \citep{stolfo2024improving}) and keep the steering weight low for MeanActDiff ($\omega = 2$), and a medium weight for SAE ($\omega = 200$),. In math reasoning tasks, we use a decaying factor of $k = 3$ or $k = 4$ with relatively high steering weights ($\omega = 600$ for Gemma-2-2b and $\omega = 400$ for Gemma-2-9b), so the coefficient decays to 1 after about 8–9 steps. We choose these hyper-parameters based on an ablation study on hyper-paramter tuning (Fig~\ref{fig:ablation}). We do not apply SAE or MeanActDiff steering to reasoning tasks with Gemma-2-9B-it, as chain-of-thought prompting is already incorporated during fine-tuning.

\subsection{Evaluation on IFEval Dataset}
From below, for simplicity, we regard $\text{SAE}_{i}$ as the i-th dimension of the trained SAE decoding matrix. Note that it's a vector $\in \mathbb{R}^{d_{sae}}$ instead of a value. The explanation for each $\text{SAE}_{i}$ could be easily accessed by~\citep{neuronpedia}. In the following steering experiments, we focus on using a single SAE feature rather than a combination of multiple features, a setting we refer to as \textit{single-feature steering}.
\paragraph{JSON cases}
We analyze the top activated SAE features between the prompt with instructions and the prompt without instructions. From algorithm 1, we found that real top activated SAE dimensions for JSON instruction following are potentially 5340, 720, and 7211, instead of the fakely top activated 1828 and 717. The removed SAE feature 1828 means the \textit{References to API requests and their parameters in code snippets}, and SAE feature 717 means the \textit{References to programming concepts and discussions around ar-
ray structures in computer science}. However, the features 5340, 720, and 7211 are all related with JSON under the explanation of gpt 4o mini by neuronpedia. If we steer with the toply activated 1828 and 717 features, it won't bring any performance improvement.

MeanActDiff steering is highly effective (97.56\% accuracy), inducing structured output even without explicit instruction.
All generations begin with ``\texttt{fastjson}'', therefore we just append this to the input and perform keyword prompting. 
We found that single-token prompting with ``\texttt{fastjson}'' performed worse than with \textbf{MeanActDiff}, so we conclude that \textbf{MeanActDiff} contains more information about json than next-token generation prompted with fastjson. In contrast, steering with SAE-derived features (e.g., $\text{SAE}_{5340}$ or $\text{SAE}_{720}$) yields variable results. Using $\text{SAE}_{5340}$ leads to 43.9\% accuracy, which is not directly equivalent to just appending a single word "JSON" at the end of the prompt (67.07\%). $\text{SAE}_{720}$ achieves low accuracy (25.61\%), suggesting partial structural influence without semantic closure. Similarly, $\text{SAE}_{7211}$ fails entirely (0\% strictly), even though it's among the top features, highlighting the importance of semantic alignment over raw activation frequency. Using SAE steering vectors is worse than using steering \textbf{MeanActDiff} vectors in this case. 
\begin{figure*}[!htb]
    \centering
    \caption{Performance plots for JSON formating, lowercase and uppercases.}
    \includegraphics[width=0.9\linewidth]{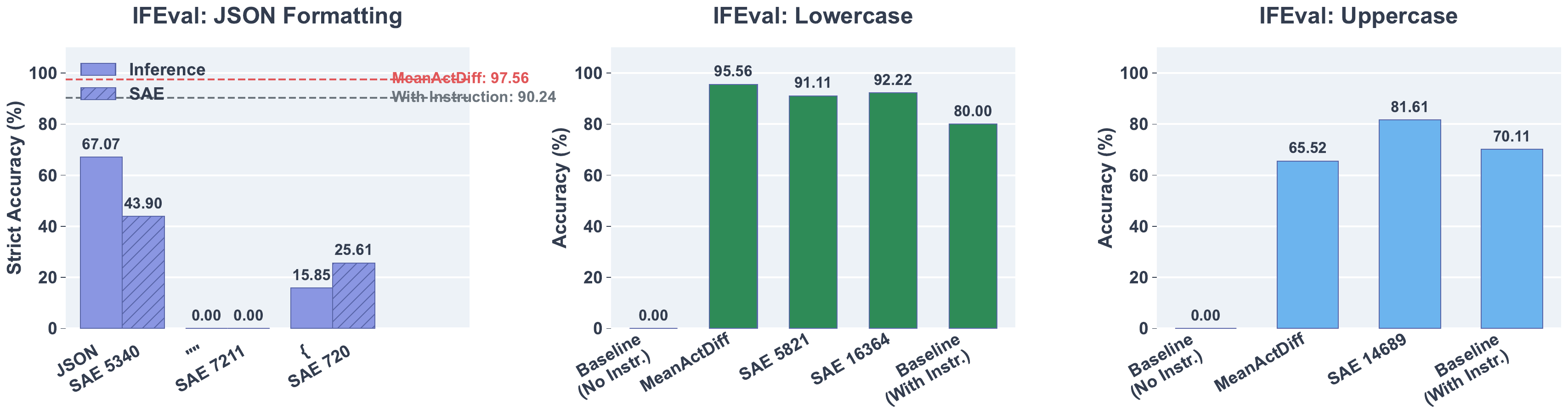}
    \label{fig:perf_plot_json}
\end{figure*}

\paragraph{Lowercase}
We compare the top activated SAE features under two settings: with instruction "Please ensure that your response is in English, and in all lower-
case letters." and without such instruction. Here, $\{1828, 5821, 16364, 7056, 717\}$ and $\{1828, 717, 6919, 15917, 7056\}$ represent the sets of top-$k$ SAE features (with $k=5$) for instruction-followed and standard prompts, respectively, averaged across the lowercase subset of the IFEval dataset. Taking the set difference, we have $\{5821, 16364\}$. These reflect both semantic domain shifts (e.g., audio) and structural sensitivities (e.g., punctuation), however they are not related with the concept lowercase at the first glance. 

% Both settings share key features such as $\text{SAE}_{1828}$ (API-related code) and $\text{SAE}_{7056}$ (health-related terminology), indicating persistent themes regardless of instruction. However, the Instruction following setting activates distinct features such as $\text{SAE}_{5821}$ (audio production terminology) and $\text{SAE}_{16364}$ (conversational punctuation structure), while the non-instruction setting instead emphasizes $\text{SAE}_{6919}$ (API error-handling code) and SAE $\text{SAE}_{15917}$(formatted Q\&A patterns). We perform constant steering with SAE feature  $\text{SAE}_{5821}$ and $\text{SAE}_{16364}$, where other top activated features are useless thus regarded.

Without instruction, it fails entirely (0.00\%), but steering with MeanActDiff restores strong performance (95.56\%), showing the model can be externally modulated to follow lowercase constraints even when instruction is absent.
Further steering with $\text{SAE}_{5821}$ and $\text{SAE}_{16364}$ yields moderately high performance (91.11\% and 92.22\%, respectively), consistent with their unique activation under the Instruction following setting. This confirms that these latent dimensions encode useful lowercase-relevant behaviors. Notably, direct instruction yields lower performance (80.00\%) than both methods, suggesting that while the model understands instructions, targeted internal steering achieves better control over formatting behavior in this case. Here \textbf{MeanActDiff} is slightly better than steering \textbf{SAE} vectors. We also recommend treating the explanations from Neuronpedia as a supplementary reference rather than a definitive source, as the interpretation precision is not guaranteed.

\paragraph{Uppercase}
Similar to lowercase redundant SAE feature analysis, we perform the same strategy to uppercase study. Common across both settings are $\text{SAE}_{1828}$ (API-related code), $\text{SAE}_{717}$ (structured technical content), and $\text{SAE}_{7056}$ (health terminology), suggesting stable relevance of these latent features regardless of instruction. Steering with these features would not direct model to generate uppercase responses.
The Instruction following condition introduces distinct high-activation features: $\text{SAE}_{14689}$ (rules and conditions in user agreements) and $\text{SAE}_{12687}$ (legal terminology), which are absent in the non-instruction setting. These may reflect more formal or rigid textual structures often associated with uppercase cases.  

Without any instruction or steering, it fails entirely (0.00\%), confirming that uppercase formatting is not default behavior. MeanActDiff achieves modest performance (65.52\%), while direct instruction improves results to 70.11\%. Interestingly, steering with instruction-free $\text{SAE}_{14689}$—the top activated SAE feature, yields the best performance (81.61\%), outperforming both MeanActDiff and instruction. This suggests $\text{SAE}_{14689}$ encodes a useful latent dimension for uppercase transformation, potentially reflecting rule-based or formal language patterns commonly found in such outputs.

\paragraph{Language cases}

\begin{table*}[!htb]
\centering
\small
\begin{tabular}{lcccccccccc}
\toprule
\textbf{Configuration} & \textbf{Ar} & \textbf{De} & \textbf{Th} & \textbf{Vi} & \textbf{Hi} & \textbf{It} & \textbf{Bn} & \textbf{Sw} & \textbf{Fa} & \textbf{Ru} \\
\midrule
Inference w/o instruction & 0 & 0 & 0 & 0 & 0 & 0 & 0 & 0 & 0 & 0 \\
MeanActDiff                & 100 & 100 & 100 & 100 & 95.83 & 100 & 100 & 42.86 & 66.67 & 100 \\
SAE                        & 75 & 100 & 100 & 100 & 66.67 & 100 & 0 & 0 & 91.67 & 100 \\
Inference w/ instruction   & 100 & 100 & 100 & 88.89 & 100 & 88.89 & 100 & 71.43 & 66.67 & 55.56 \\
\bottomrule
\end{tabular}
\caption{Gemma-2-9B-it performance (accuracy \%) on IFEval in 10 languages: Arabic (Ar), German (De), Thai (Th), Vietnamese (Vi), Hindi (Hi), Italian (It), Bengali (Bn), Swahili (Sw), Persian (Fa), and Russian (Ru).}
\label{tab:multilang-accuracy}
\end{table*}

Across 10 target languages, we observe that Instruction following reliably shifts internal SAE activations, often introducing 2 to 4 uniquely activated features per language. A small set of general-purpose SAE features—most notably $\text{SAE}_{5679}$, appear consistently across multiple languages, suggesting a shared latent pathway for instruction-based control, which is considered as the irrelevant features to language generation. In language pairs with closer syntactic or lexical ties (e.g., Hindi-Bengali or Arabic-Persian), these feature differences become less distinct, making it more challenging to isolate instruction-specific latent dimensions. This suggests that language proximity may blur the boundary between general and instruction-specific activations, and that steering in such cases may require finer-grained SAE combinations or multilingual-aware alignment strategies. In most cases,  steering SAE vectors performs on par with steering MeanActDiff.

\paragraph{Why SAE steering failed on word inclusion cases}
\begin{figure}
    \caption{Top-Activated SAE features are almost the same in word inclusion cases.}
    \centering
    \includegraphics[width=0.75\linewidth]{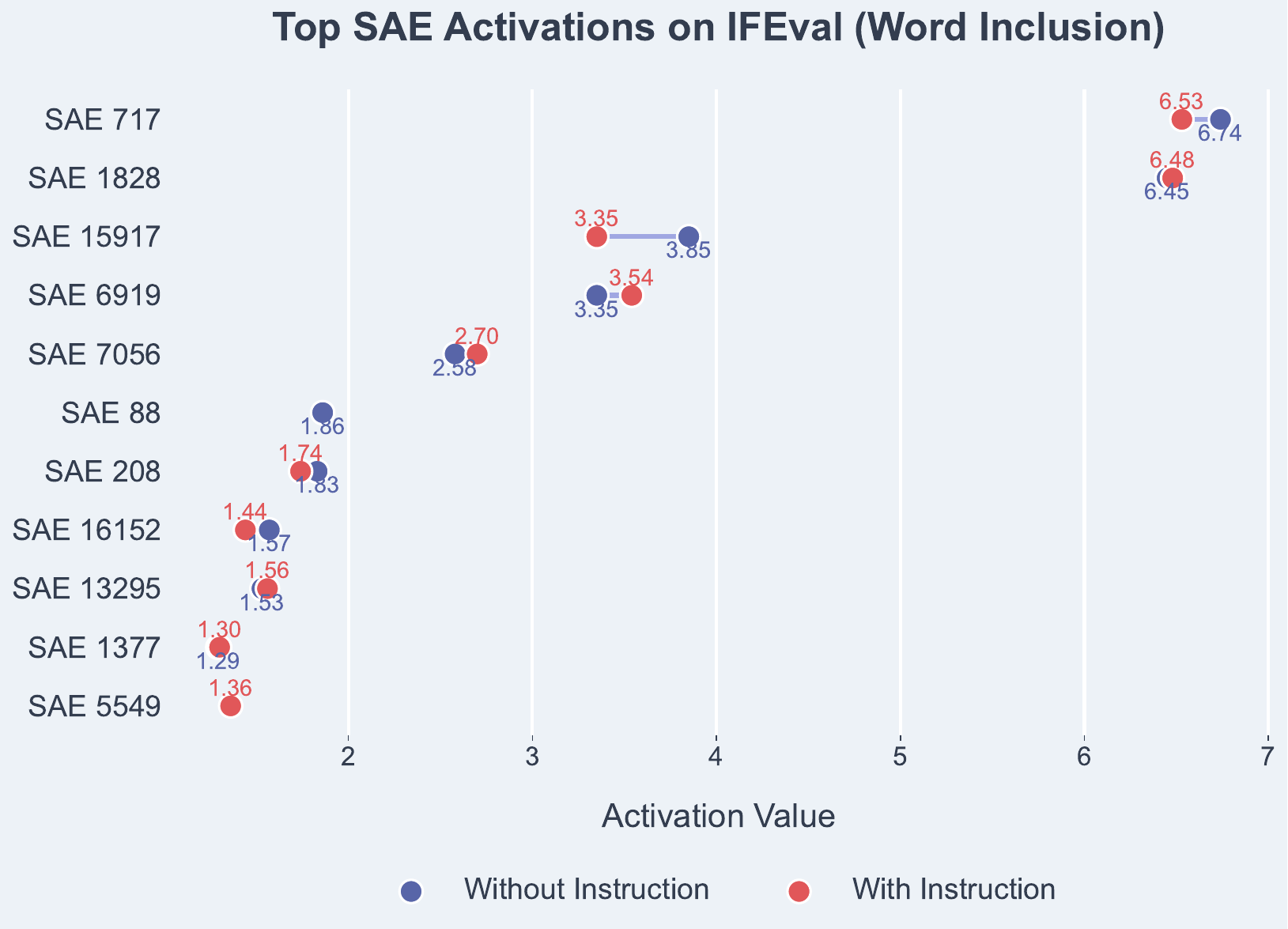}
    \label{fig:gemma-2-9b_word_inclusion}
\end{figure}
We analyze the top-10 activated SAE features for the word inclusion task, comparing models with and without instruction, as shown in Figure~\ref{fig:gemma-2-9b_word_inclusion}. Unlike previous cases (e.g., JSON formatting, response length control), the activated features are nearly identical across both conditions. Features such as $\text{SAE}_{717}$, $\text{SAE}_{1828}$, $\text{SAE}_{15917}$, and $\text{SAE}_{7056}$ appear consistently, with minimal variation in activation magnitude. No unique or dominantly activated SAE emerges that reliably correlates with the inclusion or exclusion of specific words. Even lower-ranked features such as $\text{SAE}_{88}$, $\text{SAE}_{208}$, and $\text{SAE}_{1377}$ show inconsistent behavior and are not clearly interpretable in relation to lexical content constraints. These findings suggest that word inclusion control, particularly enforcing the presence or absence of a specific keyword, is not strongly encoded along any single latent direction accessible via SAE activation. While MeanActDiff is capable of achieving this effect to some extent~\citep{stolfo2024improving}, the underlying mechanism appears to rely more on activation difference dynamics rather than discrete, interpretable SAE-based structures.

\subsection{Chain-of-thought Math Reasoning}
Table~\ref{tab:gemma-2-2b_performance} and Table~\ref{tab:gemma-2-9b_performance} present both raw and filtered accuracy across various methods. Raw accuracy is based on the last number in the generated output. However, we observe that in some cases, the correct answer appears earlier, for instance, in the second-to-last or third-to-last number. For example, given the question: \textit{"Peter has 60 dollars. 5 apples cost 5 dollars. How many 5 apples could Peter buy?"}, the model may generate: \textit{"Since 5 apples cost 5 dollars, he could buy 60 / 5 = 12 5 apples."} Although the final number is "5", the correct answer "12" appears earlier. To account for such cases, we report filtered accuracy, which scans for correct answers across the entire output.

\begin{table*}[!htb]
    \centering
    \begin{minipage}{\textwidth}
        \centering
        \resizebox{0.9\textwidth}{!}{%
            \begin{tabular}{lcc|cc|cc|cc|cc}
                \toprule
                Model Mode & CoT & SHOTS & \multicolumn{2}{c|}{GSM8K} & \multicolumn{2}{c|}{ASDIV} & \multicolumn{2}{c|}{MAWPS} & \multicolumn{2}{c}{SVAMP} \\ 
                & & & Raw & Filtered & Raw & Filtered & Raw & Filtered & Raw & Filtered \\ \midrule
                
                Gemma-2-2b  & $\times$ & 0-shot & 5.76 & 5.76 & 49.05 & 49.23 & 56.32 & 56.37 & 39.60 & 39.70 \\
                
                Gemma-2-2b + Instruction  & $\times$ & 0-shot  & 11.14 & 11.44 & 45.19 & 46.82 & 54.53 & 55.16 & 34.0 0 & 34.60 \\
                
                Gemma-2-2b + $\textit{If}$ & $\times$ & 0-shot & 15.09 & 16.38 & 43.93 & 47.63 & 54.82 & 58.50 & 38.90 & 42.30 \\

                Gemma-2-2b + MeanActDiff  & $\times$ & 0-shot  & 12.13 & 13.04 & 46.14 & 50.25 & 54.24 & 58.16 & 40.30 & 42.20 \\
                
                Gemma-2-2b + $Since$ & $\times$ & 0-shot &   16.38 & 18.43 & 53.00 & 56.25 & 66.83 & 69.69 & 42.70 & 44.20\\

                Gemma-2-2b +  $\text{SAE}_{15153}$ & $\times$ & 0-shot & 16.83 & 18.88  & 52.82 & 55.98 & 67.02 & 69.49 & 42.20 & 43.50 \\
                
                Gemma-2-2b & $\checkmark$ & 8-shots  & 27.22 & 27.30 & 60.90 & 60.90 & 79.56 & 79.66 & 48.30 & 48.50 \\
                \bottomrule
            \end{tabular}%
         }
        \caption{Gemma-2-2b: Performance on Math Reasoning Dataset}
        \label{tab:gemma-2-2b_performance}
    \end{minipage}
\end{table*}

\begin{table*}[!htb]
    \centering
    \begin{minipage}{\textwidth}
        \centering
        \resizebox{0.9\textwidth}{!}{%
            \begin{tabular}{lcc|cc|cc|cc|cc}
                \toprule
                Model Mode & CoT & SHOTS & \multicolumn{2}{c|}{GSM8K} & \multicolumn{2}{c|}{ASDIV} & \multicolumn{2}{c|}{MAWPS} & \multicolumn{2}{c}{SVAMP} \\ 
                & & & Raw & Filtered & Raw & Filtered & Raw & Filtered & Raw & Filtered \\ \midrule
                
                Gemma-2-9b  & $\times$ & 0-shot & 22.29 & 22.37 & 72.73 & 73.13 & 80.24  & 80.82 & 64.70  & 64.70 \\
                
                Gemma-2-9b + Instruction  & $\times$ & 0-shot  & 42.84 & 44.43 & 72.14  & 74.62 & 81.07  & 82.33 & 60.10  & 61.20 \\
                
                Gemma-2-9b + $\textit{First}$ & $\times$ & 0-shot &  53.90 & 58.15 & 68.98 & 73.63 & 79.23 & 83.25 & 55.60 & 57.40 \\
                
                Gemma-2-9b + MeanActDiff  & $\times$ & 0-shot  & 53.37 & 58.68  & 70.47 & 74.71 & 79.66 & 82.90 & 54.50 & 56.00 \\
                
                Gemma-2-9b + \textit{to}  & $\times$ & 0-shot  & 48.67 & 51.55 & 73.54& 76.79 & 81.60 & 85.43 & 67.50 & 68.30 \\
                
                Gemma-2-9b +  $\text{SAE}_{6782}$ & $\times$ & 0-shot & 49.13 & 52.16 & 73.59 & 76.93 & 82.08 & 85.47 & 68.60 &  70.80 \\
                
                Gemma-2-9b & $\checkmark$ & 8-shots  & 69.60 & 69.68 & 82.30 & 82.71 & 94.72 & 95.30 & 80.60 & 80.60\\

                \bottomrule
            \end{tabular}%
        }
        \caption{Gemma-2-9b: Performance on Math Reasoning Dataset}
        \label{tab:gemma-2-9b_performance}
    \end{minipage}
\end{table*}

\paragraph{Comparing MeanActDiff and Single SAE Steering}

Our experiments reveal that both MeanActDiff and single SAE feature steering are effective unsupervised methods for enhancing mathematical reasoning. As shown in Tables~\ref{tab:gemma-2-2b_performance} and \ref{tab:gemma-2-9b_performance}, both approaches consistently outperform the zero-shot baseline and often surpass simple instruction-following variants. 

MeanActDiff steering demonstrates a particularly interesting behavior: it implicitly guides the model to adopt specific lexical patterns that correlate with reasoning. For instance, on Gemma-2-2B, it consistently elicits the word \textit{If} as the initial token, achieving a filtered accuracy on GSM8K (13.04\%) nearly identical to that produced by manually prepending \textit{If} (16.38\%). A similar pattern emerges with Gemma-2-9B, where MeanActDiff steering yields 58.68\% accuracy on GSM8K by consistently generating the word \textit{First}, closely mirroring the performance of manually prompting with \textit{First} (58.15\%). These findings suggest that MeanActDiff effectively identifies and promotes a general, keyword-driven reasoning strategy.

However, steering with a single, well-chosen SAE feature proves to be a more potent and targeted strategy. Across nearly all datasets and for both models, single SAE steering outperforms MeanActDiff. For example, with Gemma-2-2B on GSM8K, $\text{SAE}_{15153}$ achieves 18.88\% filtered accuracy, significantly surpassing MeanActDiff's 13.04\%. Similarly, for Gemma-2-9B, while MeanActDiff is competitive on GSM8K, $\text{SAE}_{6782}$ shows superior performance on other datasets like SVAMP (70.80\% vs. 56.00\%). This suggests that individual SAE features can capture more nuanced and effective reasoning-related directions in activation space than the averaged vector provided by MeanActDiff.

\paragraph{Performance Ceiling and Qualitative Advantages}

Despite these successes, it is crucial to contextualize the performance of these unsupervised steering methods. The results firmly establish that \textbf{few-shot Chain-of-Thought (CoT) prompting remains the performance upper bound}. The accuracy gap between the best steering method and the 8-shot CoT baseline is significant across all tasks. For instance, Gemma-2-9B with 8-shot CoT reaches 69.68\% on GSM8K, far exceeding the 52.16\% achieved by its best-performing SAE feature. This indicates that while unsupervised steering can successfully activate reasoning capabilities, it does not yet replicate the robust, multi-step problem-solving abilities guided by explicit in-context examples.

\section{Conclusion}
We have introduced a set-subtraction-based SAE feature extractor to identify instruction-relevant features and a token-wise decaying steering mechanism to adjust steering strength dynamically during generation. Experimental results showed that in chain-of-thought reasoning tasks, both SAE and MeanActDiff steering performed similarly (Most time SAE is better) to simply prompting the model with a reasoning trigger word, such as "Since". This suggests that the model's reasoning ability may already be sufficiently activated by such prompts without requiring additional internal manipulation. In instruction-following tasks, particularly those requiring formatting control like JSON generation or lowercase responses, MeanActDiff achieved higher accuracy than SAE steering, indicating that global activation shifts capture task semantics more effectively in these settings. However, SAE steering demonstrated finer control in specific cases, such as uppercase instruction following. Steering SAE vector is on par with MeanActDiff in language case settings.
Overall, we provide a framework that could compare SAE steering more fairly with steering MeanActDiff than previous works.

\section{Limitations}
While the proposed methods showed promising results, several limitations were identified. SAE steering was effective for certain formatting tasks but failed in scenarios requiring word inclusion or exclusion, suggesting that such behaviors might rely on token-level mechanisms not captured by SAE features. In chain-of-thought reasoning, neither SAE nor MeanActDiff steering consistently outperformed simple prompting strategies, indicating limited impact on enhancing reasoning capabilities beyond surface-level interventions.
The experiments focused on single-feature SAE steering, leaving multi-feature or matrix-based steering unexplored. For example, in JSON reasoning cases, there's a better way to deal with multiple JSON-related SAE vectors to improve the performance. 
Future work could address these limitations by investigating multi-feature steering, expanding evaluations to other model architectures and tasks, and further exploring the predictive potential of SAE feature dynamics.

\section{Ethical considerations}

We use SAELens~\citep{bloom2024saetrainingcodebase} to explore SAE steering, which is currently under MIT license. We build on code released under this MIT License. Our code strictly follows their intended use as specified by the original authors (research-only). We did not employ any artifact beyond the scope of its license or stated purpose. All datasets used in this work are publicly available benchmark datasets that do not contain personally identifiable information. We reviewed the datasets and found no offensive or harmful content. We provide documentation of the datasets and artifacts used in our study. Our evaluation covers mathematical reasoning benchmarks (GSM8K, ASDIV, SVAMP, MAWPS) and instruction-following tasks from IFEval, including JSON formatting, casing (uppercase/lowercase), language control, and word inclusion. For multilingual analysis, we evaluate 10 target languages: Arabic, German, Thai, Vietnamese, Hindi, Italian, Bengali, Swahili, Persian, and Russian. These documentations specify the coverage of domains (reasoning, instruction-following), languages, and linguistic phenomena relevant to reproducibility. As our work only uses public benchmark datasets, no demographic or personally identifying information is included.

\paragraph{Usage of AI in this paper}
We used large language models (LLMs) as AI assistants to help polish and improve the writing style of the paper. They were not used for data analysis or experiment design.

\bibliography{acl_latex}
\appendix

\section{Ablation Studies}
\begin{figure*}[!htb]
    \centering
    \caption{Ablation Studies}
    \includegraphics[width=0.8\linewidth]{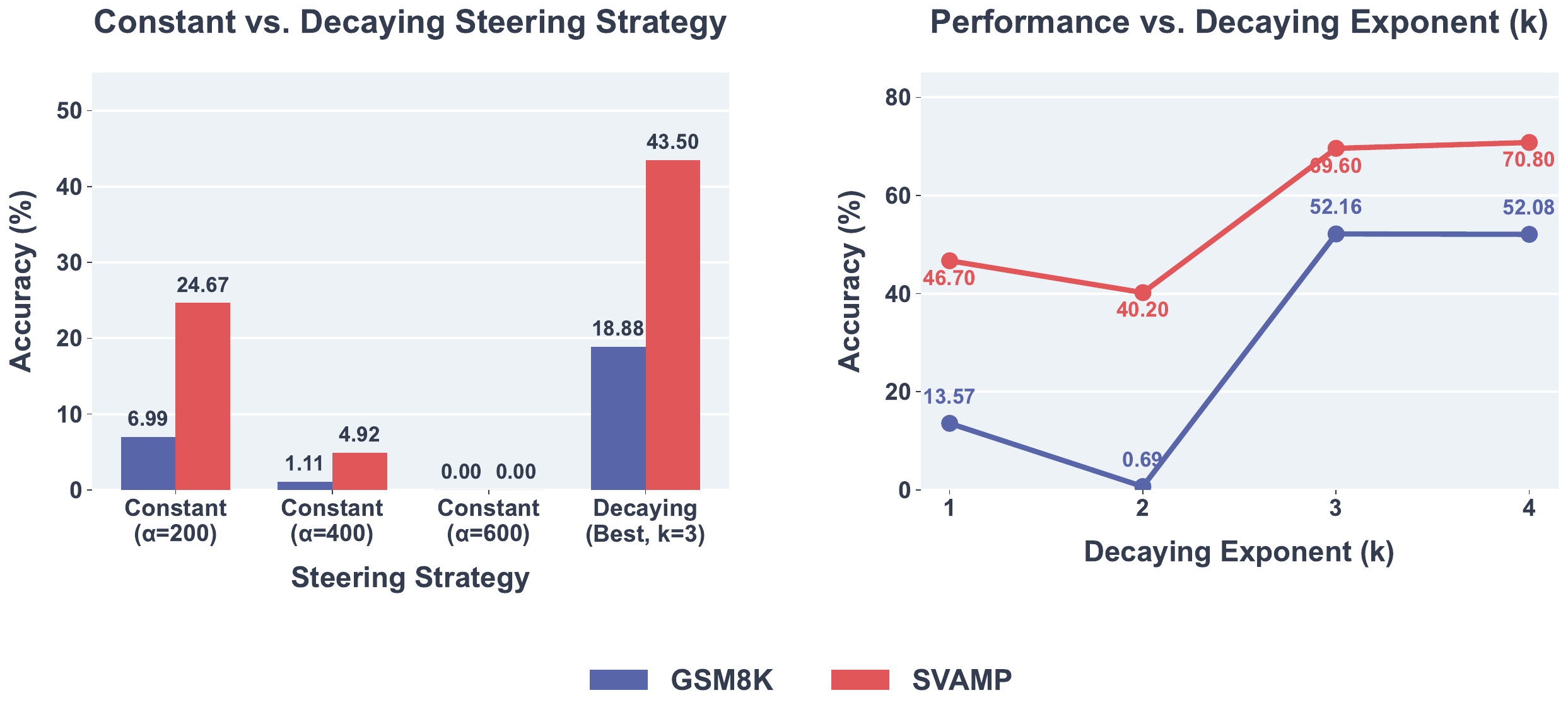}
    \label{fig:ablation}
\end{figure*}
\paragraph{Constant Steering for Math Reasoning}
As shown in Figure~\ref{fig:ablation}, we experiment with several fixed constant steering coefficients on the GSM8K and SVAMP datasets using the Gemma-2-2B model. Larger coefficients (e.g., 600) suppress the decoder heavily, resulting in performance collapse, where the generation will be "Since Since Since....". On the contrary, smaller values like 200 retain some accuracy but are still suboptimal. This is the main reason why we should apply decaying on steering coefficient.

% \begin{table}[!htb]
% \centering
% \caption{\textbf{Constant Steering vs. Decaying Steering}}
% \label{tab:constant-steering}
% \begin{tabular}{lccc}
% \toprule
% \textbf{Constant Coefficient} & GSM8K & SVAMP \\
% \midrule
% 200 & 6.99 & 24.67 \\
% 400 & 1.11 & 4.92 \\
% 600 & 0 & 0\\ \midrule
% 600 (Best Decaying) & \textbf{18.88} & \textbf{43.50} \\
% \bottomrule
% \end{tabular}
% \end{table}

\paragraph{Degree of Decaying}
% Here we briefly discuss the influence of decaying in math reasoning with SAE steering. The experiments are implemented with Gemma-2-9b model. Figure~\ref{fig:ablation} presents an ablation over the decaying exponent used in the coefficient schedule.

We keep the initial coefficient $\alpha$ to 400 in the case of math reasoning. Here $\omega$ is set to 0.5. We find that the decaying rate $k$ has a significant impact on performance. A mild decay setting ($k=1$) results in worse accuracy in math reasoning, particularly on GSM8K. Increasing the decay rate to $k=2$ also negatively affects accuracy. The best performance is observed at intermediate decay rates: $k=3$ achieves \textbf{52.16\%} accuracy on GSM8K, while $k=4$ reaches \textbf{70.80\%} on SVAMP. These results confirm that \textbf{well-tuned decaying dynamics are essential for effective steering}. In general, faster decaying of the SAE steering vector is preferred, as it provides better control over the influence of the steering signal across the generation process.

\section{Dataset Information}

\subsection{Chain-of-thought Math Reasoning}
To evaluate the effectiveness of activation steering and SAE-based interventions on multi-step reasoning tasks, we utilize several well-established benchmarks designed for mathematical reasoning. These datasets contain natural language math word problems that require logical deduction and arithmetic operations based on provided contextual information.

\paragraph{GSM8K}The GSM8K benchmark~\citep{cobbe2021training} consists of approximately 8,500 high-quality grade-school math word problems. Each question includes a detailed chain-of-thought (CoT) solution, making it a widely adopted standard for evaluating CoT prompting and alignment of reasoning in language models. We use the official test set containing 1,319 examples. Additionally, the first 500 examples from the training set are reserved for SAE feature extraction.

\paragraph{ASDIV}The Academia Sinica Diverse MWP Dataset (ASDIV)~\citep{miao-etal-2020-diverse} includes math word problems representing diverse text patterns and covering most problem types taught in elementary schools. Each problem is annotated with its problem type and corresponding grade level. The dataset contains 2,306 examples.

\paragraph{MAWPS}The MAth Word ProblemS (MAWPS) dataset~\citep{koncel-kedziorski-etal-2016-mawps} aggregates problems from various sources, including AddSub~\citep{hosseini-etal-2014-learning}, SingleOP~\citep{roy-etal-2015-reasoning}, MultiArith~\citep{roy-roth-2015-solving}, SingleEq~\citep{koncel-kedziorski-etal-2015-parsing}, SingleEq-S~\citep{kushman-etal-2014-learning}, and SingleEq-L~\citep{kushman-etal-2014-learning}, comprising a total of 2,065 examples.

\paragraph{SVAMP}The Simple Variations on Arithmetic Math word Problems (SVAMP) dataset~\citep{patel-etal-2021-nlp} is derived from MAWPS and ASDIV, providing a diagnostic dataset of simplified arithmetic word problems, consisting of 1,000 examples.
\subsection{Instruction Following}
\paragraph{IFEval Benchmark: Instruction-Following Dataset}

To measure the instruction-following capabilities of language models under precise and verifiable constraints, we adopt the IFEval benchmark~\citep{zhou2023instruction}. IFEval consists of tasks that test models' ability to adhere to specific instructions involving output format, stylistic constraints, token-level guidelines, and content patterns. These instructions are automatically evaluable through deterministic logic (e.g., regex or token-counting), ensuring reproducible evaluation.

Our experiments focus specifically on IFEval tasks relevant to steering strategies: JSON formatting, casing (uppercase and lowercase), language selection, word inclusions, and sequence length constraints. The following table \ref{tab:dataset_counts} summarizes the datasets used in this study and their corresponding counts.

\begin{table}[h]
\centering
\begin{tabular}{lc}
\hline
\textbf{Dataset / Task Type} & \textbf{Number of Samples} \\
\hline
GSM8K (Reasoning) & 1,319 \\
ASDIV (Reasoning) & 2,306 \\
SVAMP (Reasoning) & 1,000 \\
MAWPS (Reasoning) & 2,065 \\
\hline
IFEval – JSON Format & 409 \\
IFEval – Uppercase & 436 \\
IFEval – Lowercase & 454 \\
IFEval – Language Control & 378 \\
IFEval - Word inclusion & 86 \\
IFEval – Length Constraint & 1,297 \\
\hline
\textbf{Total (Used)} & \textbf{9,750} \\
\hline
\end{tabular}
\caption{Overview of datasets used in our study and its statistics}
\label{tab:dataset_counts}
\end{table}

\section{More Algorithms}

\begin{algorithm}[!htb]
\caption{Differentially Activated Feature Extraction: Cumulative Activations}
\label{alg:diff_activated_features_2}
\KwIn{LLM $\mathcal{F}$, SAE Encoder $\mathcal{S}_{l,enc}$, layer index $l$, datasets $\mathcal{T}, \mathcal{T}^*$, integer $K$}
\KwOut{Set of differentially activated SAE feature indices}

\SetKwBlock{Begin}{Initialize:}{}
\Begin{
    $\mathbf{S}, \mathbf{S}^* \gets \{\}, \{\}$\;
    $\mathbf{z}_{cum}$, $\mathbf{z}^*_{cum} \gets 0, 0$
    % \text{Frequency counters for top-K indices}
}

\For{$i \gets 0$ \KwTo $K-1$}{
    % 1. Get the full sequence of activations for each instance
    $\mathbf{H}_{l}, \mathbf{H}^*_{l} \gets \mathcal{A}_l(\mathcal{F}(\mathcal{T}_i)), \mathcal{A}_l(\mathcal{F}(\mathcal{T}^*_i))$ 

    % 2. Encode the entire sequence of activations
    $\mathbf{Z}_{l}, \mathbf{Z}^*_{l} \gets \mathcal{S}_{l,enc}(\mathbf{H}_{l}), \mathcal{S}_{l,enc}(\mathbf{H}^*_{l})$ 
    % \text{Matrix of shape (seq\_len, d\_sae)} \;
    
    % 3. Select the sparse code vector for the final token
    $\mathbf{z}_{l, \text{last}}, \mathbf{z}^*_{l, \text{last}} \gets \mathbf{Z}_{l}[-1], \gets \mathbf{Z}^*_{l}[-1]$\;

    % Update frequency counters
     $\mathbf{z}_{cum} \gets \mathbf{z}_{cum} + \mathbf{z}_{l, \text{last}}$\;
            $\mathbf{z}^*_{cum} \gets \mathbf{z}^*_{cum} + \mathbf{z}^*_{l, \text{last}}$\;
        }
        % Find Top-K from the final summed vectors
        $\text{indices}, \text{indices}^* \gets \text{TopK}(\mathbf{z}_{cum}, K)$, $\text{TopK}(\mathbf{z}^*_{cum}, K)$\;
        
        \Return $\text{indices}^* \setminus \text{indices}$
\end{algorithm}

An alternative, cumulative activation approach for identifying discriminative features is detailed in Algorithm~\ref{alg:diff_activated_features_2}. In contrast to the frequency-based method, this algorithm aggregates the SAE feature activations before identifying the most salient ones. It begins by initializing two zero vectors, $\mathbf{z}_{cum}$ and $\mathbf{z}^*_{cum}$, to serve as accumulators. Within the loop over the datasets, after the final-token latent vectors ($\mathbf{z}_{l, \text{last}}$ and $\mathbf{z}^*_{l, \text{last}}$) are extracted for each instance pair, they are summed into their respective cumulative vectors. This process continues for all $K$ instances. Only after the loop is complete are the top-$K$ feature indices extracted from the final summed vectors $\mathbf{z}_{cum}$ and $\mathbf{z}^*_{cum}$. This method hypotheses that the importance of a feature is captured not only by how often it appears but also by its average activation magnitude across the entire dataset. The final set of discriminative features is again found by taking the set difference between the indices from the augmented and standard datasets.

\section{SAE Feature Distribution}
We report the SAE feature distribution, and then show the tables of the top activated features for the chain-of-thought-math reasoning cases, as well as for some instruction following cases. The Figures are shown from Figure 6 to Figure 17.

\begin{figure*}[!htb]
    \centering

    \begin{minipage}[t]{0.3\textwidth}
        \centering
        \includegraphics[width=\textwidth]{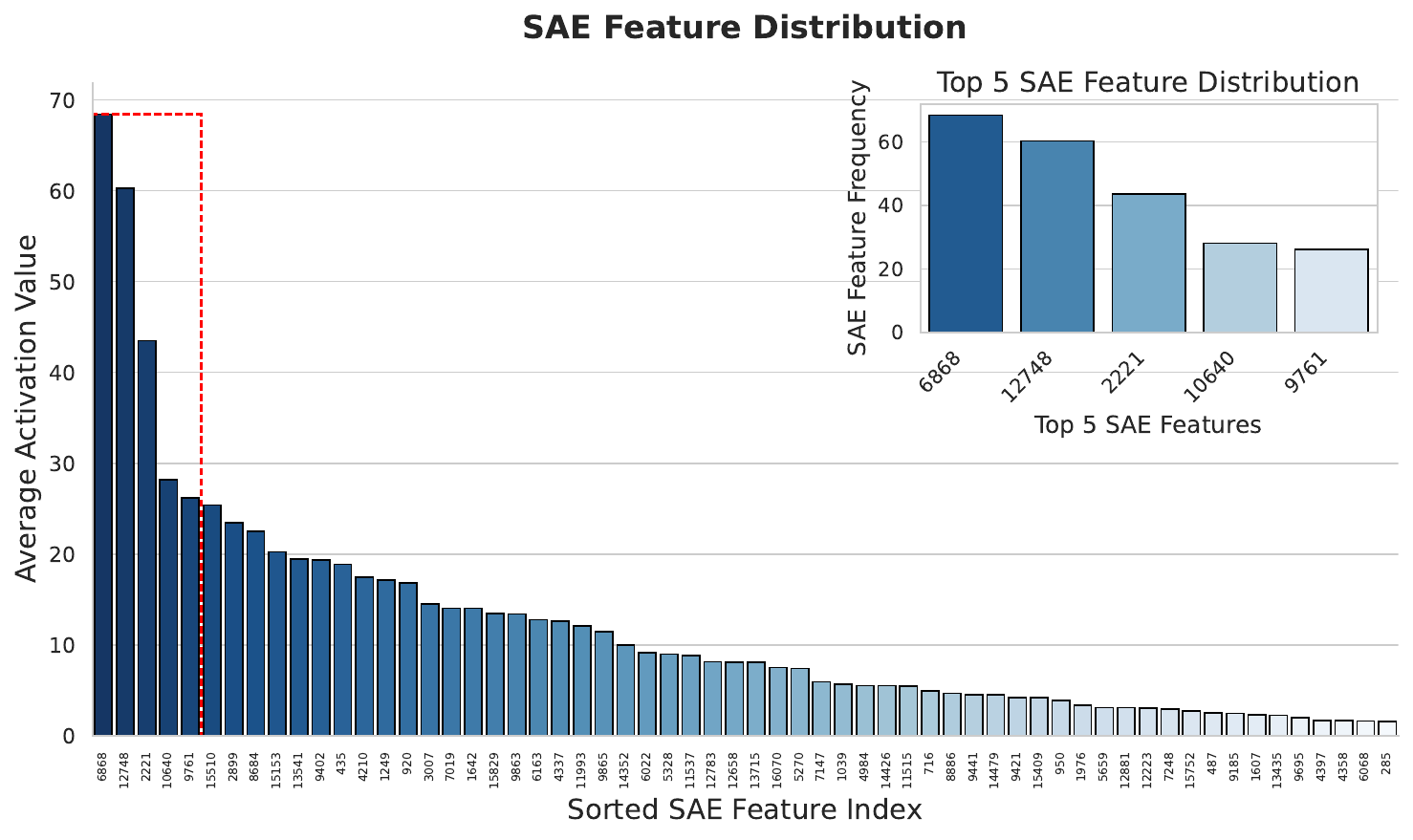}
        \caption{Gemma-2-2b (w/o CoT)}
        \label{fig:gemma-2-2b-wo-cot}
    \end{minipage}
    \begin{minipage}[t]{0.3\textwidth}
        \centering
        \includegraphics[width=\textwidth]{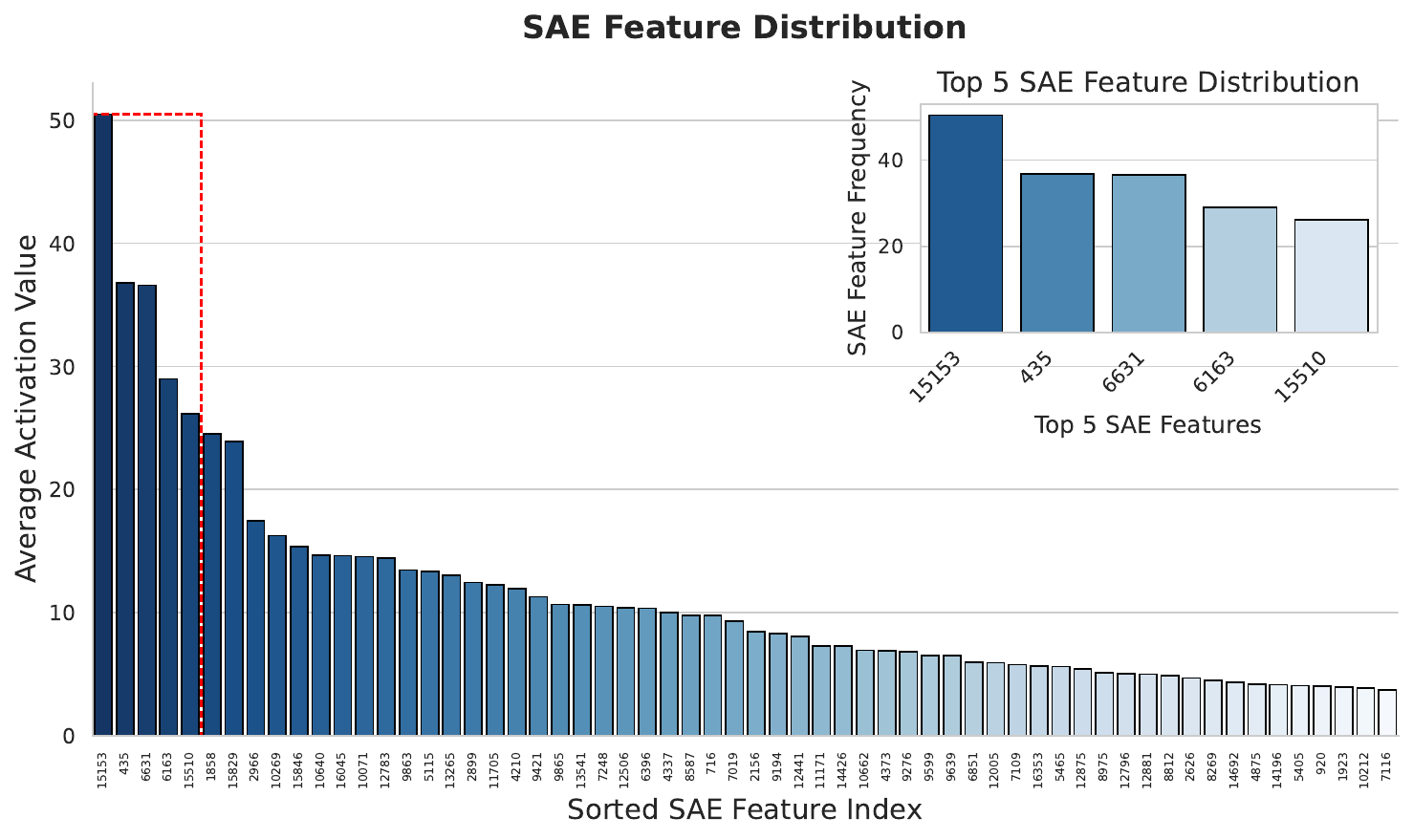}
        \caption{Gemma-2-2b (w/ CoT)}
        \label{fig:gemma-2-2b-w-cot}
    \end{minipage}
    \begin{minipage}[t]{0.3\textwidth}
        \centering
        \includegraphics[width=\textwidth]{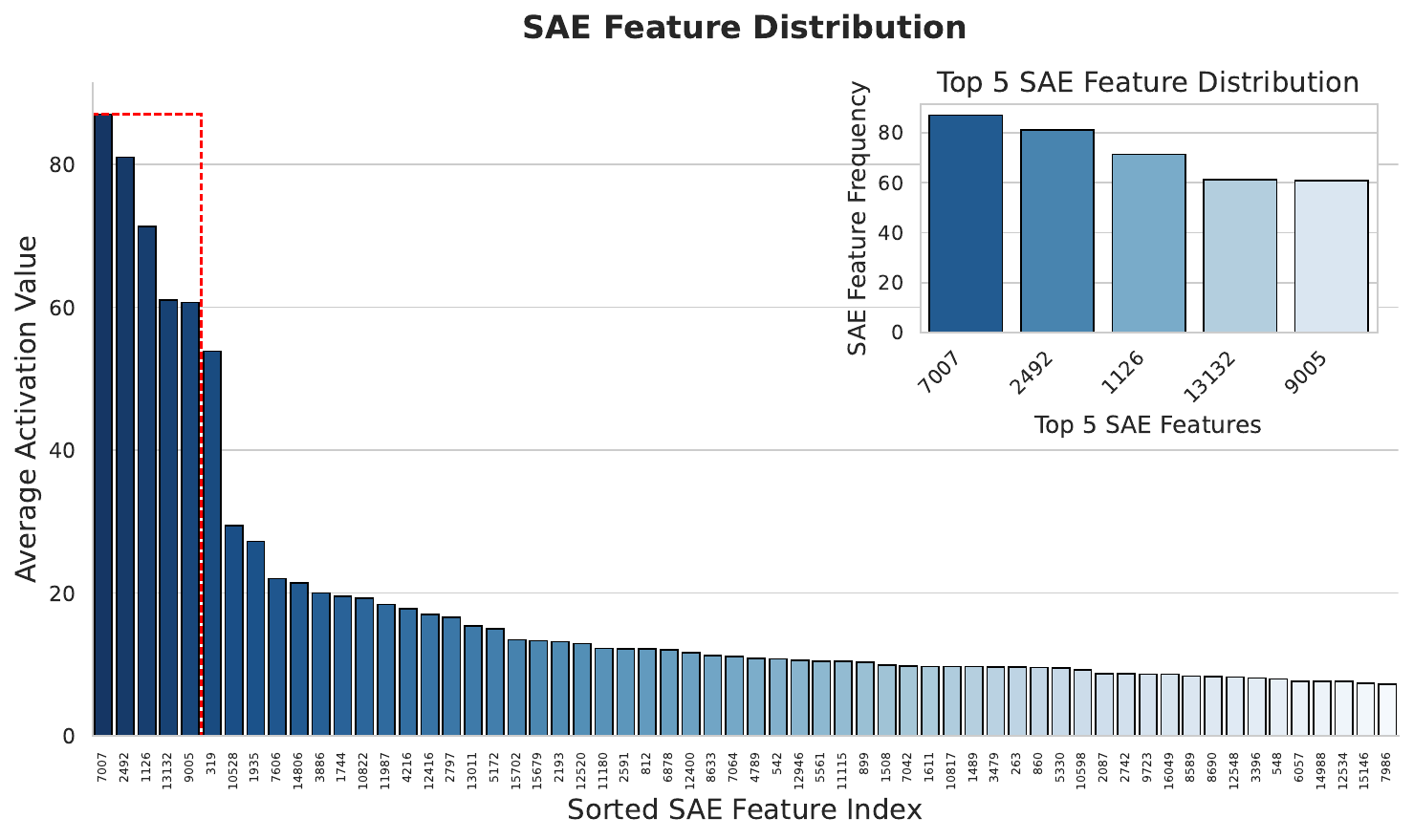}
        \caption{Gemma-2-9b (w/o CoT)}
        \label{fig:gemma-2-9b-wo-cot}
    \end{minipage}

    \begin{minipage}[t]{0.3\textwidth}
        \centering
        \includegraphics[width=\textwidth]{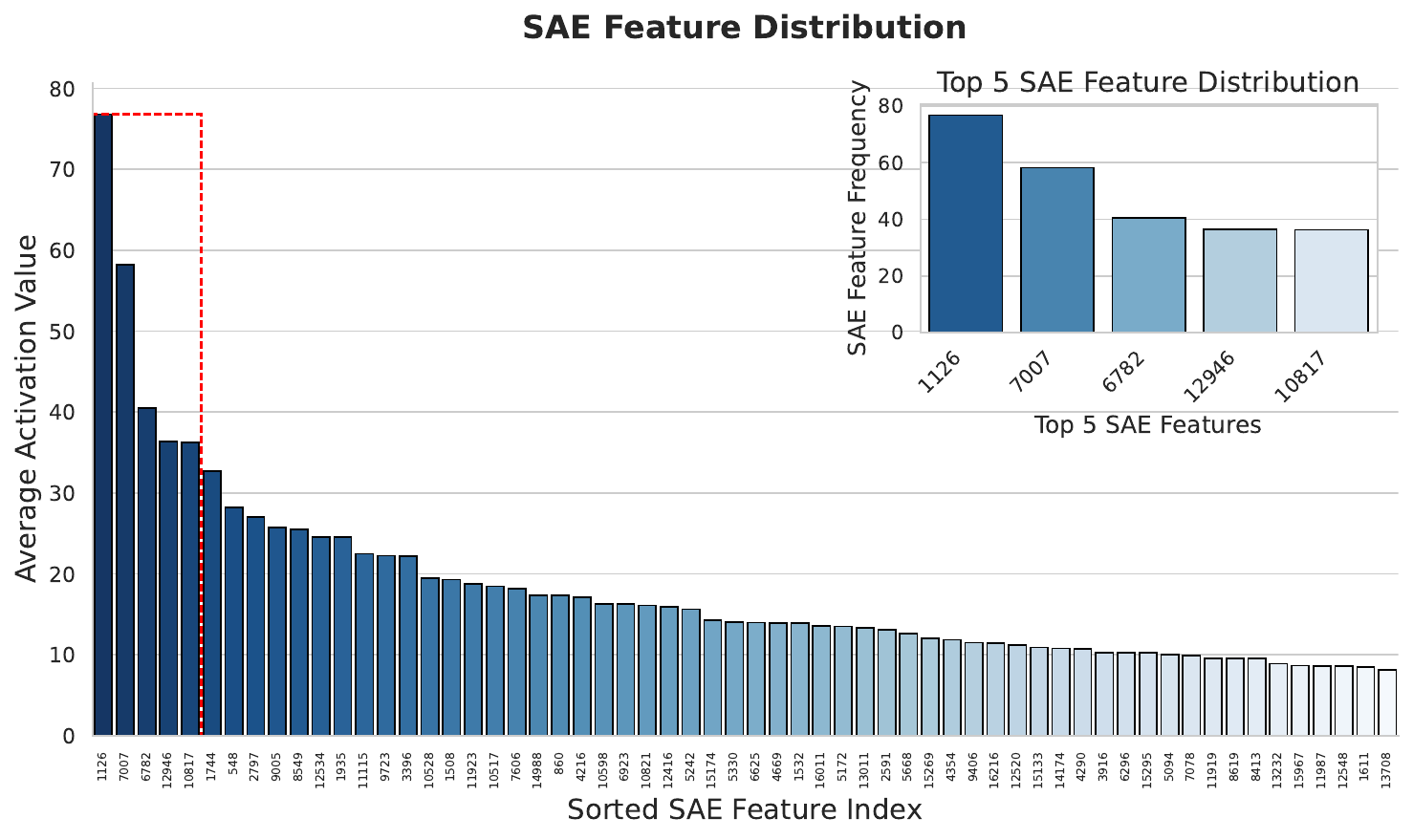}
        \caption{Gemma-2-9b (w/ CoT)}
        \label{fig:gemma-2-9b-w-cot}
    \end{minipage}
    \begin{minipage}[t]{0.3\textwidth}
        \centering
        \includegraphics[width=\textwidth]{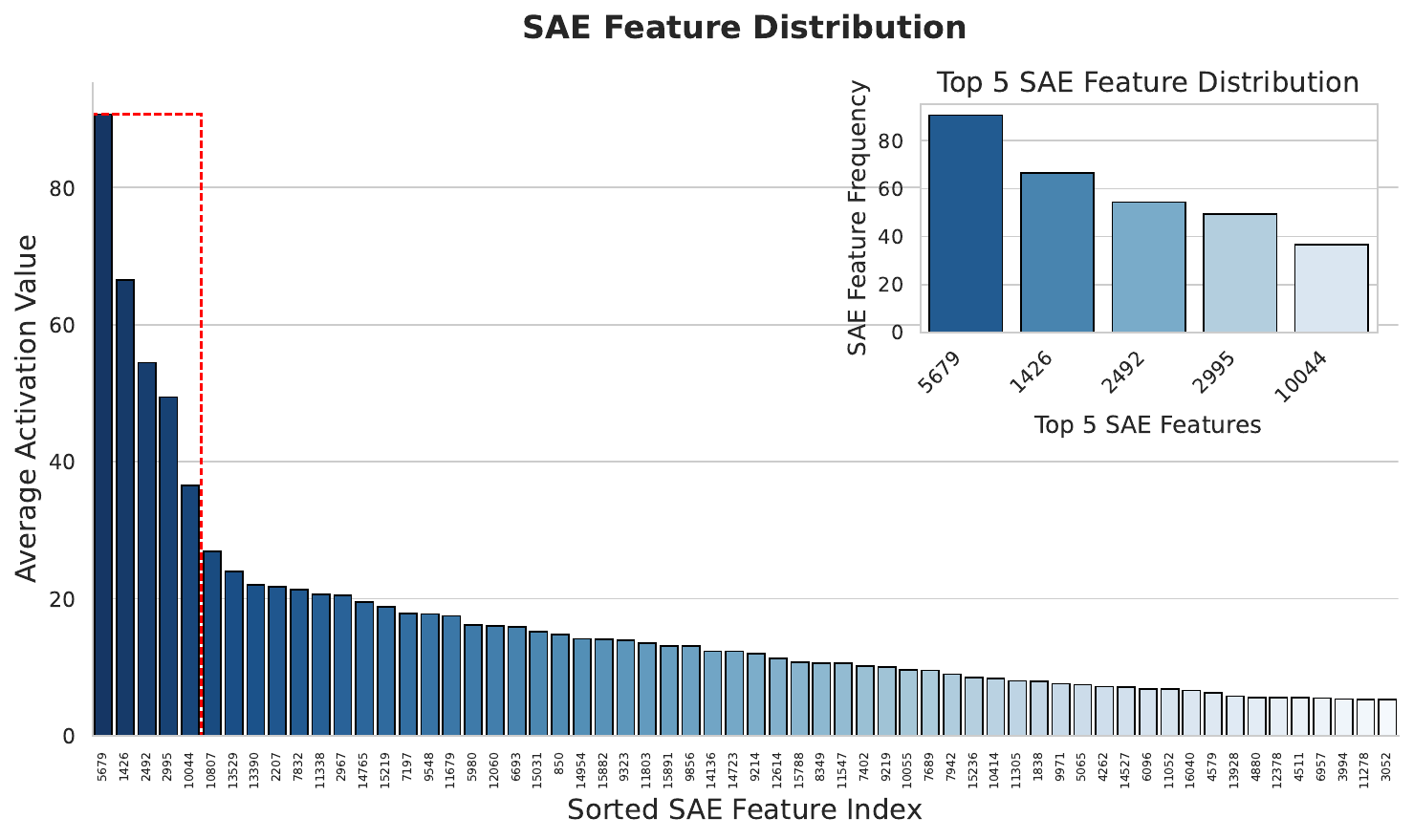}
        \caption{Gemma-2-9-it (w/o CoT)}
        \label{fig:gemma-2-9-it-wo-cot}
    \end{minipage}
    \begin{minipage}[t]{0.3\textwidth}
        \centering
        \includegraphics[width=\textwidth]{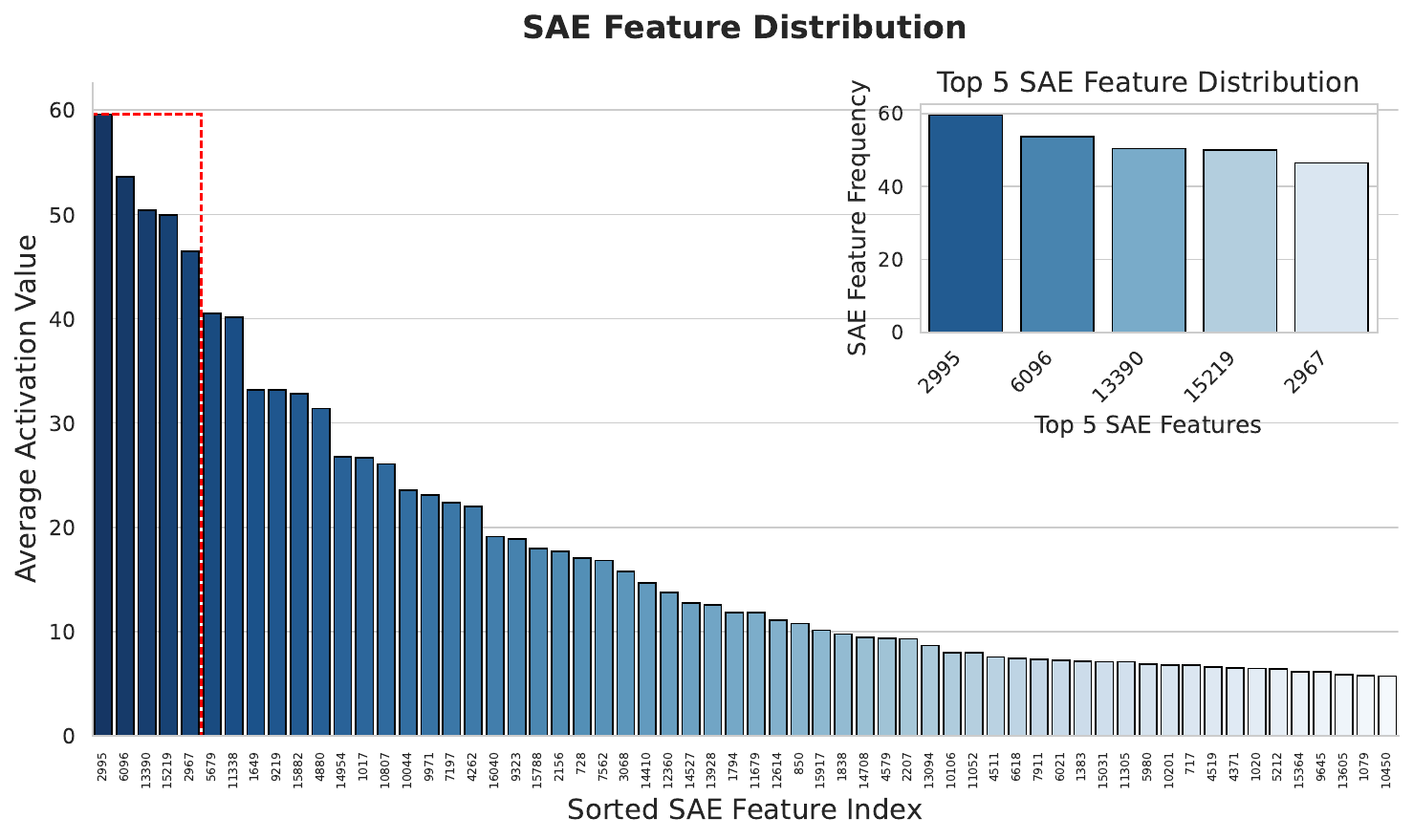}
        \caption{Gemma-2-9-it (w/ CoT)}
        \label{fig:gemma-2-9-it-w-cot}
    \end{minipage}

    \begin{minipage}[t]{0.3\textwidth}
        \centering
        \includegraphics[width=\textwidth]{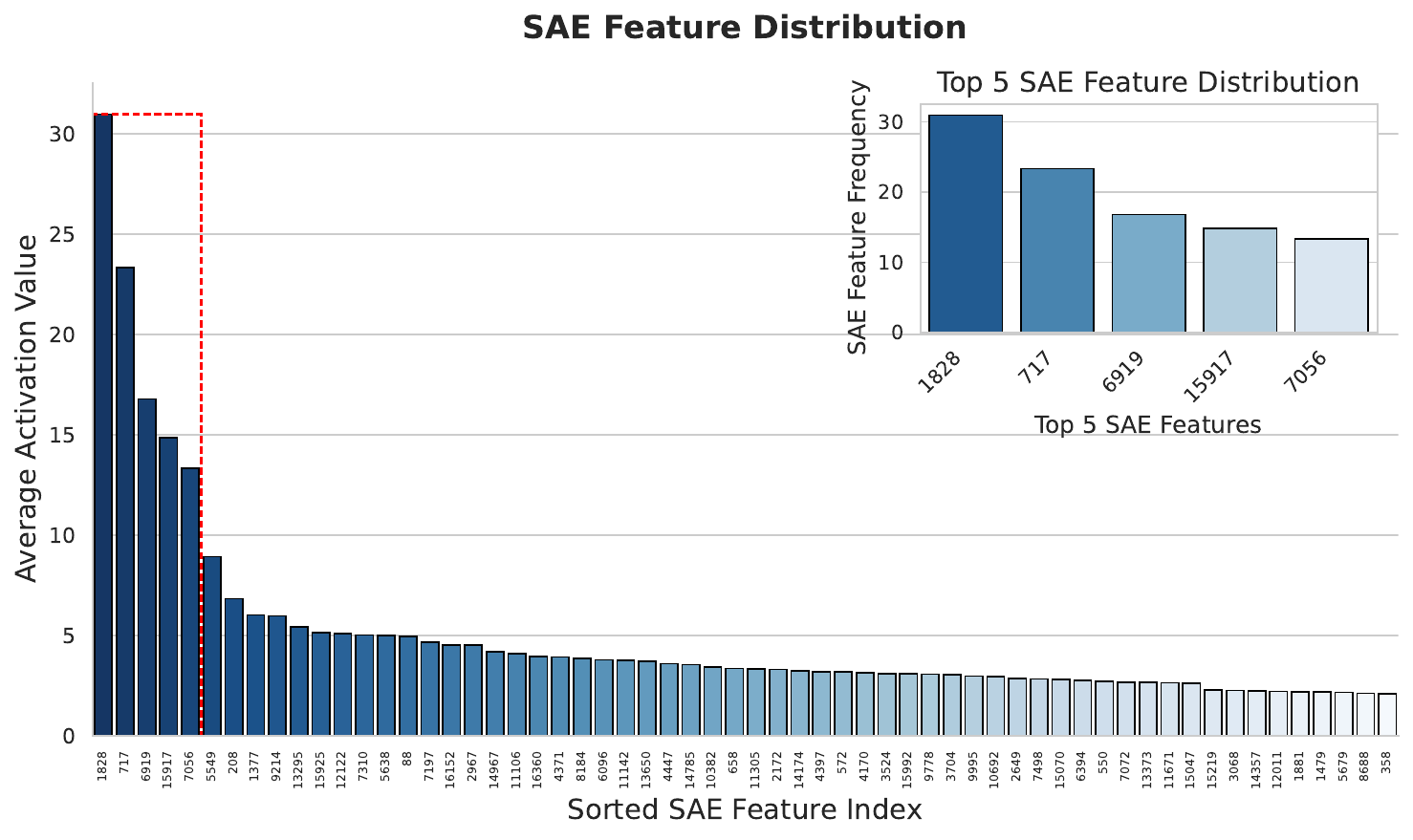}
        \caption{Gemma-2-9b-it (w/o instr.) on JSON}
        \label{fig:gemma-2-9b-it-wo-instr-json}
    \end{minipage}
    \begin{minipage}[t]{0.3\textwidth}
        \centering
        \includegraphics[width=\textwidth]{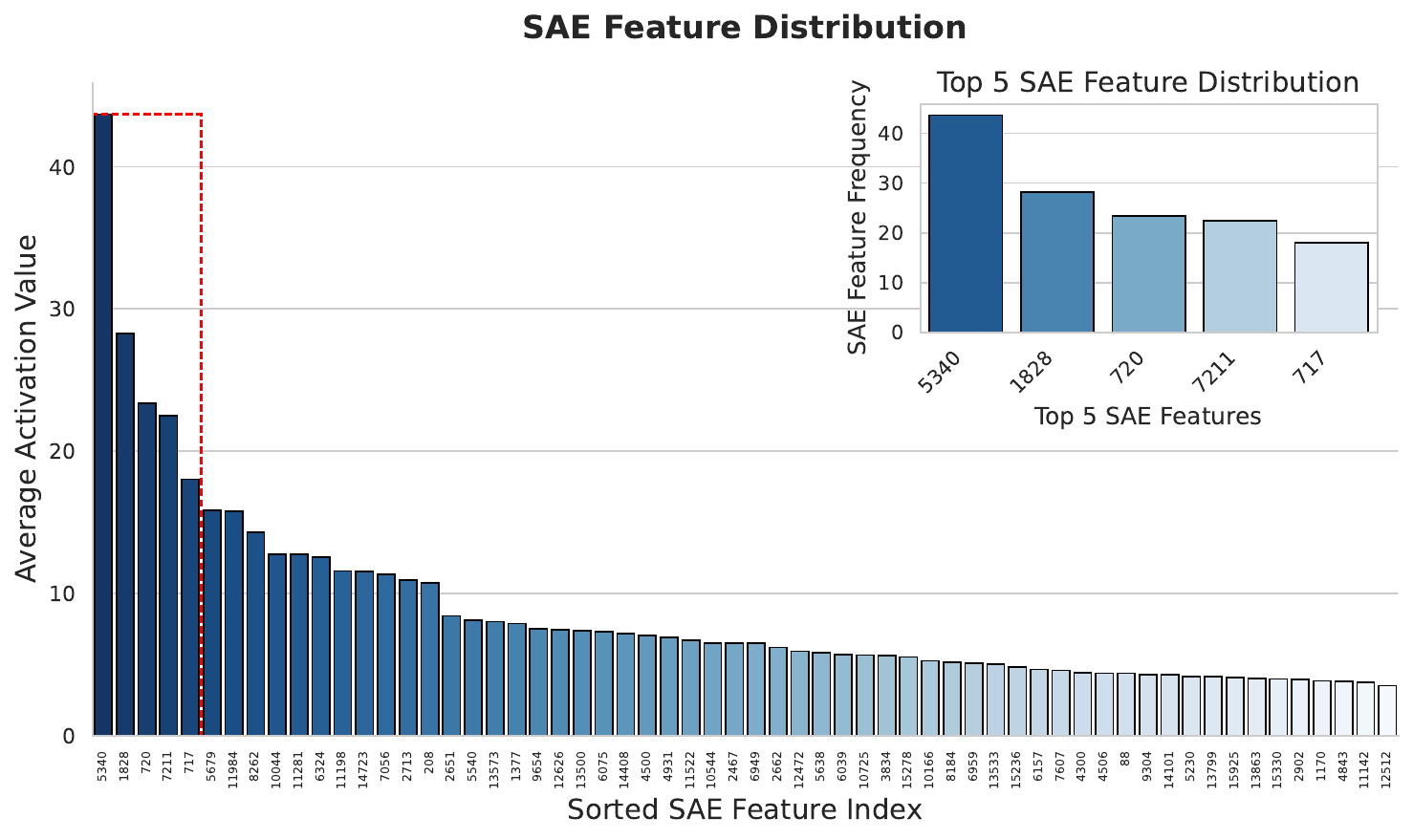}
        \caption{Gemma-2-9b-it (w/ instr.) on JSON}
        \label{fig:gemma-2-9b-it-w-instr-json}
    \end{minipage}
    \begin{minipage}[t]{0.3\textwidth}
        \centering
        \includegraphics[width=\textwidth]{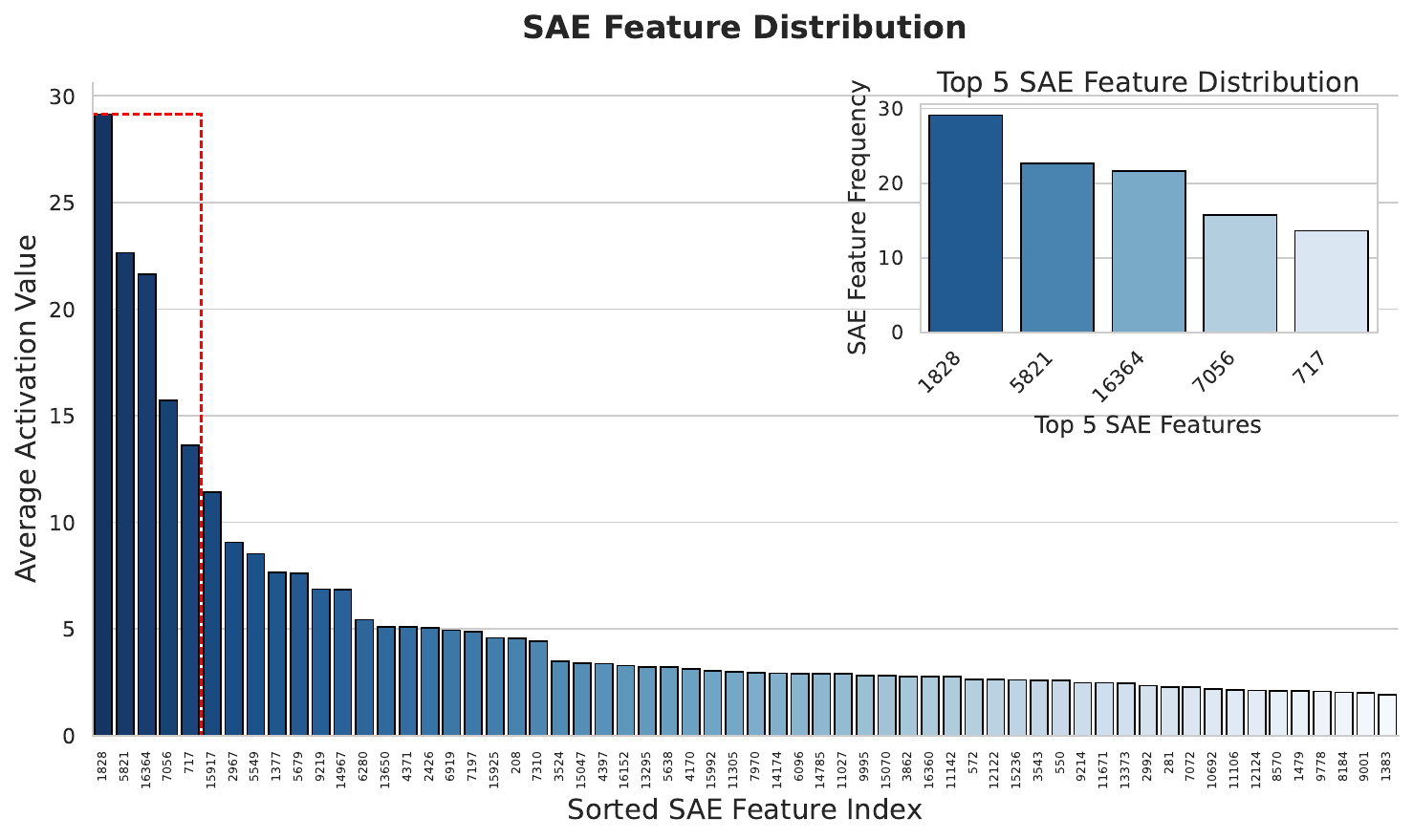}
        \caption{Gemma-2-9b-it (w/o instr.) on Lowercase}
        \label{fig:gemma-2-9b-it-wo-instr-lowercase}
    \end{minipage}

    \begin{minipage}[t]{0.3\textwidth}
        \centering
        \includegraphics[width=\textwidth]{imgs/appendix/lowercase/SAE_gemma-2-9b-it_barplot_5_COT.pdf}
        \caption{Gemma-2-9b-it (w/ instr.) on Lowercase}
        \label{fig:gemma-2-9b-it-w-instr-lowercase}
    \end{minipage}
    \begin{minipage}[t]{0.3\textwidth}
        \centering
        \includegraphics[width=\textwidth]{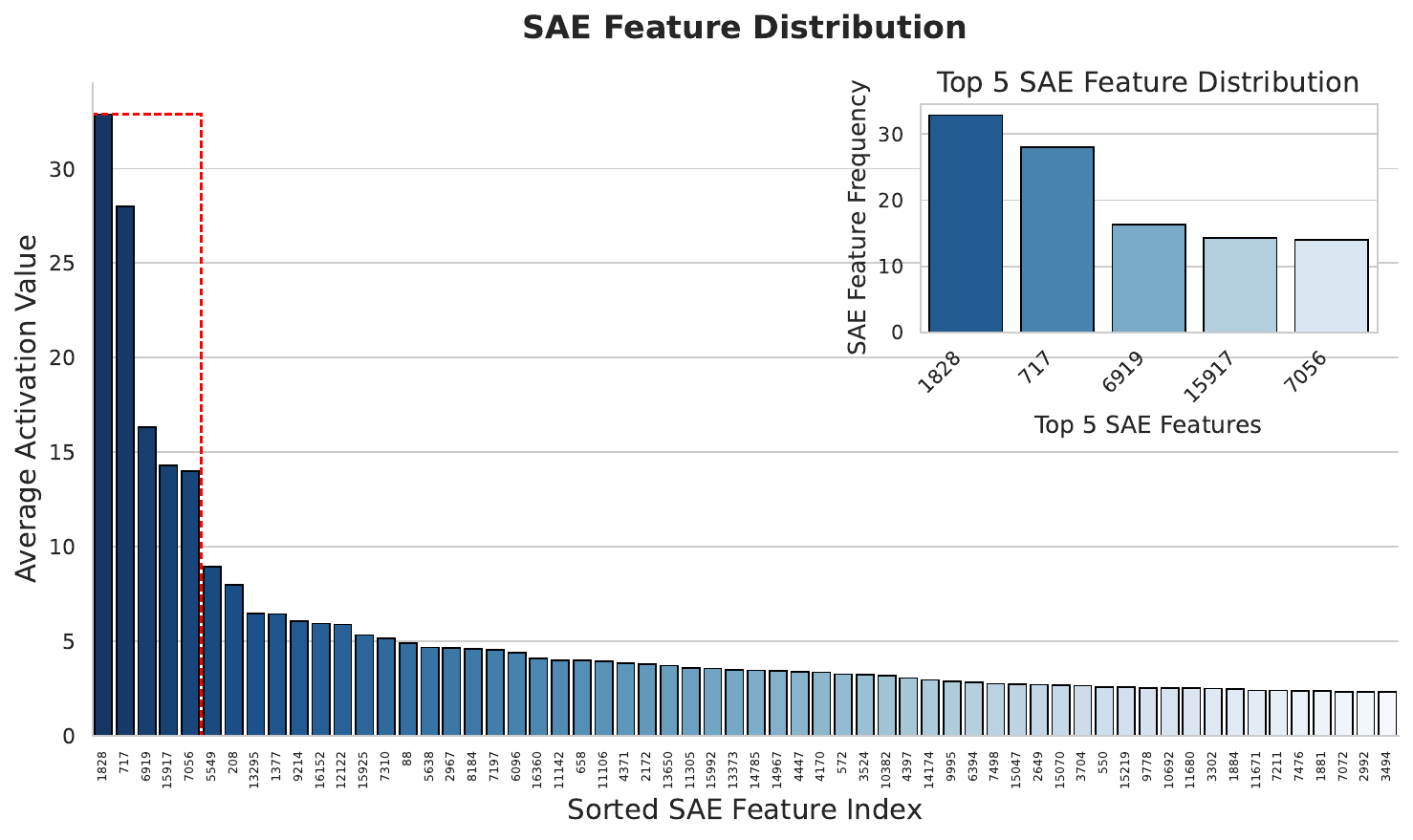}
        \caption{Gemma-2-9b-it (w/o instr.) on Uppercase}
        \label{fig:gemma-2-9b-it-wo-instr-uppercase}
    \end{minipage}
    \begin{minipage}[t]{0.3\textwidth}
        \centering
        \includegraphics[width=\textwidth]{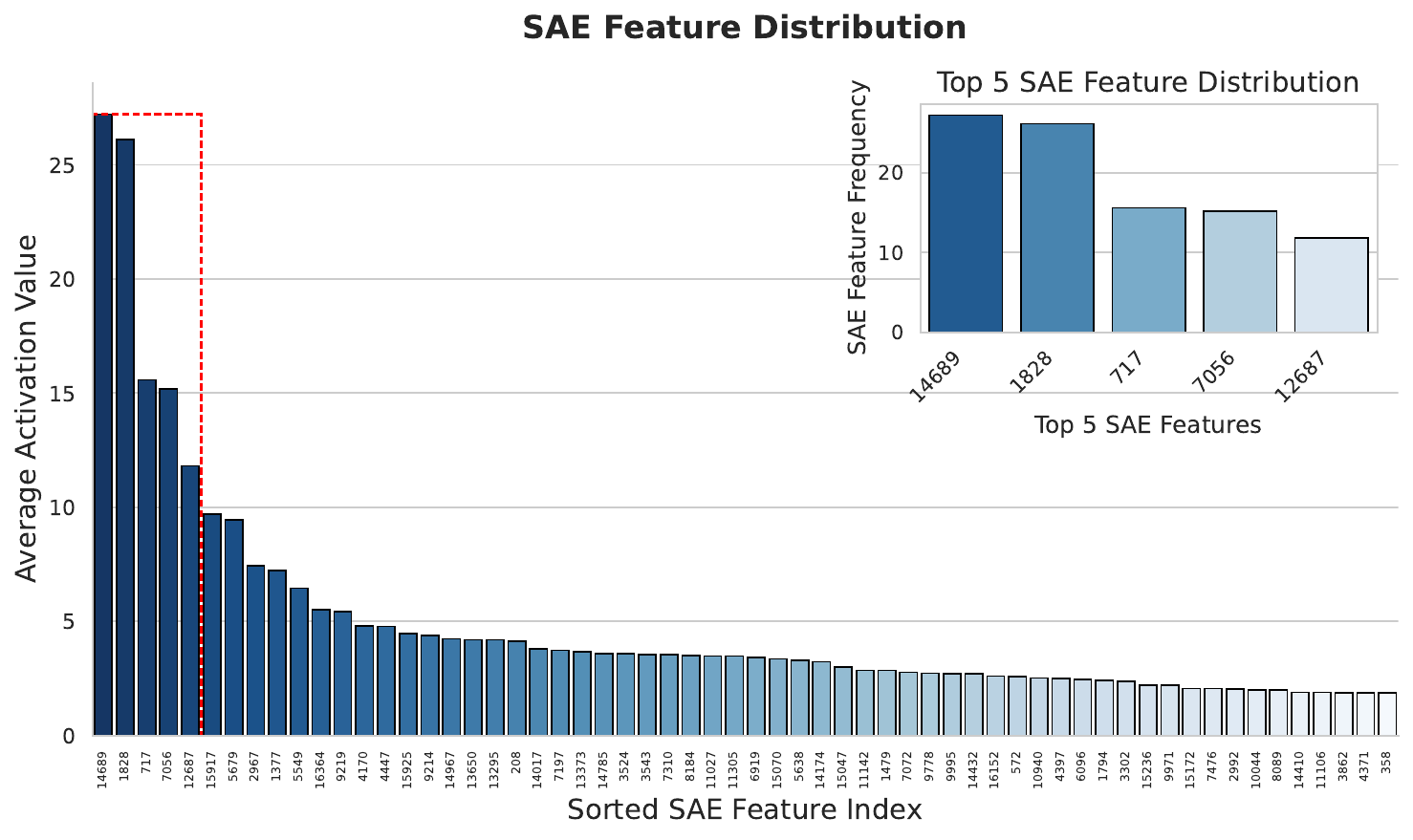}
        \caption{Gemma-2-9b-it (w/ instr.) on Uppercase}
        \label{fig:gemma-2-9b-it-w-instr-uppercase}
    \end{minipage}

    \label{fig:sae-barplot-comparison}
\end{figure*}

\section{SAE Feature and Their Explanations}
\subsection{Chain-of-thought math reasoning}
We typically set $K = 10$. Results for the Gemma-2 model family, including Gemma-2-2B, Gemma-2-9B, and Gemma-2-9B-it, are presented in the main findings. The corresponding sparse autoencoders are pre-trained in Gemma-Scope. The tables below show the top 5 most activated SAE features (out of the top 10). We define $F^{k}_{\text{cot}}$ as the set of $k$ most active features on Chain-of-Thought (CoT) inputs, and $F^{k}_{\text{std}}$ as the set of $k$ most active features on standard inputs without CoT.

\begin{table*}[!htb]
    \centering
    % 使用 tabularx 环境，并设置总宽度为 \linewidth
    % @{} c: Rank 列, 居中
    % c: Index 列, 居中
    % X: Feature Description 列, 自动换行, 两端对齐
    % c: Activation 列, 居中
    \begin{tabularx}{\linewidth}{@{} c c X c @{}}
        \toprule
        \textbf{Rank} & \textbf{Index} & \textbf{Feature Description} & \textbf{Activation} \\
        \midrule % 使用 \midrule 分隔表头和数据
        1 & 6631 & the beginning of a text or important markers in a document & 63.25 \\ 
        2 & 15153 & mathematical expressions and relationships & 35.24 \\ 
        3 & 1858 & punctuation marks and sentence endings & 33.51 \\ 
        4 & 1642 & numeric values and their relationships in a technical context & 27.37 \\ 
        5 & 2899 & specific instructions or steps related to a process & 24.51 \\
        \bottomrule
    \end{tabularx}
    \caption{Top 5 activated SAE features for Gemma-2-2B with Chain-of-Thought}
    \label{tab:gemma2b_table_w_cot_2B}
\end{table*}

\begin{table*}[!htb]
    \centering
    % 使用 tabularx 环境，并设置总宽度为 \linewidth
    % @{} c: Rank 列, 居中
    % c: Index 列, 居中
    % X: Feature Description 列, 自动换行, 两端对齐
    % c: Activation 列, 居中
    \begin{tabularx}{\linewidth}{@{} c c X c @{}}
        \toprule
        \textbf{Rank} & \textbf{Index} & \textbf{Feature Description} & \textbf{Activation} \\
        \midrule % 使用 \midrule 分隔表头和数据
                1 & 1126 & numerical data and references in structured formats & 73.33 \\ 
                2 & 1744 & mathematical terms and expressions &  35.54\\ 
                3 & 1877 & technical or programming-related terms and functions & 35.48\\ 
                4 & 1935 & sentences with numerical data and measurements related to various subjects & 30.99\\ 
                5 & 11115 & dates and structured data representations & 24.01 \\
                \bottomrule
            \end{tabularx}%
        \caption{Top 5 activated SAE features for Gemma-2-9B without Chain-of-Thought}
        \label{tab:gemma9b_table_wo_cot_2B}
\end{table*}

\begin{table*}[!htb]
    \centering
    % 使用 tabularx 环境，并设置总宽度为 \linewidth
    % @{} c: Rank 列, 居中
    % c: Index 列, 居中
    % X: Feature Description 列, 自动换行, 两端对齐
    % c: Activation 列, 居中
    \begin{tabularx}{\linewidth}{@{} c c X c @{}}
        \toprule
        \textbf{Rank} & \textbf{Index} & \textbf{Feature Description} & \textbf{Activation} \\
        \midrule % 使用 \midrule 分隔表头和数据
                1 & 7007 & terms related to automotive diagnostics and their connection methods &  98.91 \\ 
                2 & 1126 & numerical data and references in structured formats &  88.26 \\ 
                3 & 9005 & terms and phrases related to healthcare interventions and studies & 71.25 \\ 
                4 & 6782 & technical terms and processes related to scientific algorithms and methodologies & 47.57 \\ 
                5 & 12946 & phrases indicating reasons or justifications in legal or analytical contexts & 28.53\\ 
                \bottomrule
            \end{tabularx}%

        \caption{Top 5 activated SAE features for Gemma-2-9B with Chain-of-Thought}
        \label{tab:gemma9b_table_w_cot_2B}
\end{table*}

\begin{table*}[!htb]
    \centering
    % 使用 tabularx 环境，并设置总宽度为 \linewidth
    % @{} c: Rank 列, 居中
    % c: Index 列, 居中
    % X: Feature Description 列, 自动换行, 两端对齐
    % c: Activation 列, 居中
    \begin{tabularx}{\linewidth}{@{} c c X c @{}}
        \toprule
        \textbf{Rank} & \textbf{Index} & \textbf{Feature Description} & \textbf{Activation} \\
        \midrule % 使用 \midrule 分隔表头和数据
                1 & 5679 & references to specific data science tools and technologies & 90.71 \\ 
                2 & 1426 & phrases with statistical or probabilistic references &  66.50 \\ 
                3 & 2492 & phrases related to processes and outcomes in negotiations or transactions & 54.44 \\ 
                4 & 2995 &different types of data structures and their attributes & 49.43 \\ 
                5 & 10044 & references to statistical analysis and methods in research contexts & 36.55 \\
                \bottomrule
            \end{tabularx}%
        \caption{Top 5 activated SAE features for Gemma-2-2B-it without Chain-of-Thought}
        \label{tab:gemma9bit_table_wo_cot_2B}
\end{table*}

\begin{table*}[!htb]
    \centering
    % 使用 tabularx 环境，并设置总宽度为 \linewidth
    % @{} c: Rank 列, 居中
    % c: Index 列, 居中
    % X: Feature Description 列, 自动换行, 两端对齐
    % c: Activation 列, 居中
    \begin{tabularx}{\linewidth}{@{} c c X c @{}}
        \toprule
        \textbf{Rank} & \textbf{Index} & \textbf{Feature Description} & \textbf{Activation} \\
        \midrule % 使用 \midrule 分隔表头和数据
                1 & 2995 & different types of data structures and their attributes & 59.61 \\ 
                2 & 6096 & phrases and contexts that indicate differing interpretations or meanings &  53.60 \\ 
                3 & 13390 & phrases and terms related to legal analysis and reasoning in court cases & 50.40 \\ 
                4 & 15219 & elements of mathematical expressions and functions & 49.95 \\ 
                5 & 2967 & technical terms and coding references related to programming and web development & 46.46 \\
                \bottomrule
            \end{tabularx}%
        \caption{Top 5 activated SAE features for Gemma-2-9B-it without Chain-of-Thought}
        \label{tab:gemma9bit_table_w_cot_2B}
\end{table*}

 Further, we define the $\textit{SAE-cardinality} (SC)$ as the cardinality of the difference between the set $F^{k}_{cot}$ and the set $F^{k}_{std}$:
\begin{equation}
    \text{SC} = \left| F^{k}_{cot} \setminus F^{k}_{std} \right|
\end{equation}
For Gemma-2-2b, $$F^{5}_{std} = \{6868, 12748, 2221, 10640, 9761\}$$ $$F^{5}_{cot} = \{6631, 15153, 1858, 1642, 2899\}$$ $ \Rightarrow \text{Gemma2-2b-SC}_{5} = \left| F^{5}_{cot} \right| = 5$.

For Gemma-2-9b, $$F^{5}_{std} = \{1126, 1744, 1877, 1935, 11115\}$$  $$F^{5}_{cot} = \{7007, 1126, 9005, 6782, 12946\}$$ $ \Rightarrow \text{Gemma2-2b-SC}_{5} = \left| \{7007, 9005, 6782, 12946\}\right| = 4$.

For Gemma-2-9-it, $$F^{5}_{std} = \{5679, 1426, 2492, 2995, 10044\}$$ $$F^{5}_{cot} = \{2995, 6096, 13390, 15219, 2967\}$$ $ \Rightarrow \text{Gemma2-2b-SC}_{5} = \left| \{6096, 13390, 15219, 2967\}\right| = 4$.

We use this metric to avoid steering redundant or meaningless SAE features. For example, in Gemma-2-9B models, we remove feature 1126, \text{described as} \textit{data in structured formats}. The model instead identifies its two most active features as \textbf{terms related to automotive diagnostics and their connection methods} and \textbf{phrases indicating reasons}, which align with Chain-of-Thought prompting. (In further experiments, we show that the top activated SAE feature 7007 does not perform as well as SAE feature 6782.)
For the instruction-tuned Gemma-2-9B model, feature 2995, \text{described as} \textit{different types of data structures}, is removed. This is similar to the top active but unhelpful feature 1126 in the base Gemma-2-9B model. As a result, the model now surfaces top active features \textit{interpreted as} \textbf{phrases indicate differing interpretations} and \textbf{phrases related to legal analysis and reasoning}, both of which suggest the model is engaging in Chain-of-Thought reasoning.

Generally we define $F^{k}_{inst}$ as the set of k features active the most on Instruction Followed inputs, and  $F^{k}_{std}$ as the set of k features active the most on standard inputs without instructions.  The steering experiments are performed under zero-shot setting, except for the length constraint case where instructions are appended to the input.
\subsection{JSON Format}
\begin{table*}[!htb]
    \centering
    % 使用 tabularx 环境，并设置总宽度为 \linewidth
    % @{} c: Rank 列, 居中
    % c: Index 列, 居中
    % X: Feature Description 列, 自动换行, 两端对齐
    % c: Activation 列, 居中
    \begin{tabularx}{\linewidth}{@{} c c X c @{}}
        \toprule
        \textbf{Rank} & \textbf{Index} & \textbf{Feature Description} & \textbf{Activation} \\
        \midrule % 使用 \midrule 分隔表头和数据
            1 & 1828 & References to API requests and their parameters in code snippets & 30.98  \\ 
            2 & 717 & Structured elements within technical documentation &  23.32\\ 
            3 & 6919 & Code snippets related to API data retrieval and error handling & 16.78 \\ 
            4 & 15917 & Formatted questions and responses & 14.83 \\ 
            5 & 7056 & Scientific concepts and terminologies related to health and wellness & 13.31 \\
            \bottomrule
        \end{tabularx}
        \caption{Top 5 activated SAE features for Gemma-2-9B-it without instruction (JSON)}
        \label{tab:gemma2b_table_wo_ins_json}

\end{table*}

\begin{table*}[!htb]
    \centering
    % 使用 tabularx 环境，并设置总宽度为 \linewidth
    % @{} c: Rank 列, 居中
    % c: Index 列, 居中
    % X: Feature Description 列, 自动换行, 两端对齐
    % c: Activation 列, 居中
    \begin{tabularx}{\linewidth}{@{} c c X c @{}}
        \toprule
        \textbf{Rank} & \textbf{Index} & \textbf{Feature Description} & \textbf{Activation} \\
        \midrule % 使用 \midrule 分隔表头和数据
            1 & 5340 & JavaScript and JSON-related code structures and functions & 43.69\\ 
            2 & 1828 & References to API requests and their parameters in code snippets & 28.26\\ 
            3 & 720 & Keywords and syntax related to JavaScript code structure, particularly involving function definitions and JSON processing & 23.35 \\ 
            4 & 7211 & Structured elements within technical documentation & 22.48\\ 
            5 & 717 & References to programming concepts and discussions around array structures in computer science & 18.03 \\
            \bottomrule
        \end{tabularx}
        \caption{Top 5 activated SAE features for Gemma-2-9B-it with instruction (JSON)}
        \label{tab:gemma2b_table_w_ins_json}
\end{table*}

\paragraph{Feature-Level Activation Analysis}

To understand the effect of Instruction following behavior in JSON generation, we analyze the top activated SAE features between the Instruction following setting and the baseline (no instruction).  Following the cumulative extraction method, we average SAE activations across the dataset, then compute the Top-K (with $k=5$) most salient dimensions.

Tables~\ref{tab:gemma2b_table_wo_ins_json} and~\ref{tab:gemma2b_table_w_ins_json} show the top 5 activated SAE indices and their associated feature descriptions. From these, we perform set subtraction $F^{k}_{\text{inst}} \setminus F^{k}_{\text{std}} = \{5340, 720, 7211\}$, which yields the features most uniquely activated by instruction-followed inputs. Specifically, SAE features 5340, 720, and 7211 emerge as discriminative latents, associated with structured JSON-related code, function syntax, and programmatic structure. Feature 1828 and feature 717 are not useful since we could see these two features in other cases below as well.

\subsection{Lowercase}

\begin{table*}[!htb]
    \centering
    % 使用 tabularx 环境，并设置总宽度为 \linewidth
    % @{} c: Rank 列, 居中
    % c: Index 列, 居中
    % X: Feature Description 列, 自动换行, 两端对齐
    % c: Activation 列, 居中
    \begin{tabularx}{\linewidth}{@{} c c X c @{}}
        \toprule
        \textbf{Rank} & \textbf{Index} & \textbf{Feature Description} & \textbf{Activation} \\
        \midrule % 使用 \midrule 分隔表头和数据
            1 & 1828 & references to API requests and their parameters in code snippets & 34.05  \\ 
            2 & 717 & structured elements within technical documentation
 &  26.05\\ 
            3 & 6919 & code snippets related to API data retrieval and error handling & 16.73 \\ 
            4 & 15917 & formatted questions and responses & 15.21 \\ 
            5 & 7056 & scientific concepts and terminologies related to health and wellness & 13.31 \\
            \bottomrule
        \end{tabularx}
        \caption{Top 5 activated SAE features for Gemma-2-9B-it without instruction (Lowercase)}
        \label{tab:gemma2b_table_wo_ins_lowercase}
\end{table*}

\begin{table*}[!htb]
    \centering
    % 使用 tabularx 环境，并设置总宽度为 \linewidth
    % @{} c: Rank 列, 居中
    % c: Index 列, 居中
    % X: Feature Description 列, 自动换行, 两端对齐
    % c: Activation 列, 居中
    \begin{tabularx}{\linewidth}{@{} c c X c @{}}
        \toprule
        \textbf{Rank} & \textbf{Index} & \textbf{Feature Description} & \textbf{Activation} \\
        \midrule % 使用 \midrule 分隔表头和数据
            1 & 1828 & references to API requests and their parameters in code snippets & 29.14  \\ 
            2 & 5821 &terms and concepts related to audio production and technology &  22.65\\ 
            3 & 16364 & symbols and punctuation marks that indicate structure in conversation, such as question marks, quotes, and parentheses & 21.64 \\ 
            4 & 7056 & scientific concepts and terminologies related to health and wellness & 15.71 \\ 
            5 & 717 & structured elements within technical documentation & 13.61 \\
            \bottomrule
        \end{tabularx}
        \caption{Top 5 activated SAE features for Gemma-2-9B-it with instruction (Lowercase)}
        \label{tab:gemma2b_table_w_ins_lowercase}
\end{table*}

\paragraph{Feature-Level Activation Analysis}

To assess the model's behavior when handling lowercase outputs in Instruction following tasks, we compare the top activated SAE features under two settings: with instruction "Please ensure that your response is in English, and in all lower-
case letters." and without such instruction. Here, $F^{k}_{\text{inst}} = \{1828, 5821, 16364, 7056, 717\}$ and $F^{k}_{\text{std}} = \{1828, 717, 6919, 15917, 7056\}$ represent the sets of top-$k$ SAE features (with $k=5$) for instruction-followed and standard prompts, respectively, averaged across the lowercase subset of the IFEval dataset. Taking the set difference $F^{k}_{\text{inst}} \setminus F^{k}_{\text{std}} = \{5821, 16364\}$, we identify latent dimensions specifically enhanced by Instruction following prompts. These reflect both semantic domain shifts (e.g., audio) and structural sensitivities (e.g., punctuation), however they are not related with the concept lowercase at the first glance. 

As shown in Tables~\ref{tab:gemma2b_table_wo_ins_lowercase} and~\ref{tab:gemma2b_table_w_ins_lowercase}, both settings share key features such as $\text{SAE}_{1828}$ (API-related code) and $\text{SAE}_{7056}$ (health-related terminology), indicating persistent themes regardless of instruction. However, the Instruction following setting activates distinct features such as $\text{SAE}_{5821}$ (audio production terminology) and $\text{SAE}_{16364}$ (conversational punctuation structure), while the non-instruction setting instead emphasizes $\text{SAE}_{6919}$ (API error-handling code) and SAE $\text{SAE}_{15917}$(formatted Q\&A patterns). We perform constant steering with SAE feature  $\text{SAE}_{5821}$ and $\text{SAE}_{16364}$, where other top activated features are useless thus regarded.

\subsection{Uppercase}

\begin{table*}[!htb]
    \centering
    % 使用 tabularx 环境，并设置总宽度为 \linewidth
    % @{} c: Rank 列, 居中
    % c: Index 列, 居中
    % X: Feature Description 列, 自动换行, 两端对齐
    % c: Activation 列, 居中
    \begin{tabularx}{\linewidth}{@{} c c X c @{}}
        \toprule
        \textbf{Rank} & \textbf{Index} & \textbf{Feature Description} & \textbf{Activation} \\
        \midrule % 使用 \midrule 分隔表头和数据
            1 & 1828 & references to API requests and their parameters in code snippets & 32.87  \\ 
            2 & 717 & structured elements within technical documentation
 &  28.01\\ 
            3 & 6919 & code snippets related to API data retrieval and error handling & 16.32 \\ 
            4 & 15917 & formatted questions and responses & 14.31 \\ 
            5 & 7056 & scientific concepts and terminologies related to health and wellness & 14.00 \\
            \bottomrule
        \end{tabularx}
        \caption{Top 5 activated SAE features for Gemma-2-9B-it without instruction (Uppercase)}
        \label{tab:gemma2b_table_wo_ins_uppercase}
\end{table*}

\begin{table*}[!htb]
    \centering
    % 使用 tabularx 环境，并设置总宽度为 \linewidth
    % @{} c: Rank 列, 居中
    % c: Index 列, 居中
    % X: Feature Description 列, 自动换行, 两端对齐
    % c: Activation 列, 居中
    \begin{tabularx}{\linewidth}{@{} c c X c @{}}
        \toprule
        \textbf{Rank} & \textbf{Index} & \textbf{Feature Description} & \textbf{Activation} \\
        \midrule % 使用 \midrule 分隔表头和数据
            1 & 14689 & phrases indicating rules, conditions, or restrictions regarding user agreements and obligations & 27.21  \\ 
            2 & 1828 & references to API requests and their parameters in code snippets &  26.11\\ 
            3 & 717 & structured elements within technical documentation & 15.58 \\ 
            4 & 7056 & scientific concepts and terminologies related to health and wellness & 15.19 \\ 
            5 & 12687 & specific legal terms and references within a judicial context & 11.81 \\
            \bottomrule
        \end{tabularx}
        \caption{Top 5 activated SAE features for Gemma-2-9B-it with instruction (Uppercase)}
        \label{tab:gemma2b_table_w_ins_uppercase}
\end{table*}

\paragraph{Feature-Level Activation Analysis}

Tables~\ref{tab:gemma2b_table_wo_ins_uppercase} and~\ref{tab:gemma2b_table_w_ins_uppercase} report the most salient SAE features in each case. Common across both settings are $\text{SAE}_{1828}$ (API-related code), $\text{SAE}_{717}$ (structured technical content), and $\text{SAE}_{7056}$ (health terminology), suggesting stable relevance of these latent features regardless of instruction. Steering with these features would not direct model to generate uppercase responses.
The Instruction following condition introduces distinct high-activation features: $\text{SAE}_{14689}$ (rules and conditions in user agreements) and $\text{SAE}_{12687}$ (legal terminology), which are absent in the non-instruction setting. These may reflect more formal or rigid textual structures often associated with uppercase use cases.

\subsection{Language Cases}
\begin{table*}[!htb]
    \centering
    \resizebox{\textwidth}{!}{
    \begin{tabular}{c|ccccc}
        \toprule
        \midrule
        \textbf{Language} & \textbf{SAE Rank = 1} & \textbf{SAE Rank = 2} & \textbf{SAE Rank = 3} & \textbf{SAE Rank = 4} & \textbf{SAE Rank = 5} \\
        \midrule
        Arabic (w/o inst.) & 1828 & 6919 & 717 & 8688 & 5549\\
        Arabic (w/ inst.)  & 6356 & 5679 & 717 & 1828 & 7056 \\ \midrule
        Bengali (w/o inst.)  & 1828 & 6919 & 717 & 7197 & 7056 \\
        Bengali (w/ inst.)   & 6874 & 1828 &5679 &31 & 6098 \\ \midrule
        German (w/o inst.)  & 1828 & 717 & 7056 & 6919 & 13295 \\
        German (w/ inst.)   & 12251 & 1828 & 717 & 5679 & 15917 \\ \midrule
        Persian (w/o inst.)  & 1828 & 717 & 13650 & 1479 & 10564 \\
        Persian (w/ inst.)   & 16267 & 1828 & 13341 & 717 & 5679 \\ \midrule
        Hindi (w/o inst.)  & 1828 & 6919 & 717 & 15917 & 208 \\
        Hindi (w/ inst.)   & 1828 & 7056 & 5679 & 31 & 6098 \\\midrule
        Italian (w/o inst.)  & 1828 & 7056 & 8985 & 5549 & 6919 \\
        Italian (w/ inst.)   & 1828 & 4427 & 5679 & 7056 & 14174 \\\midrule
        Swahili (w/o inst.)  & 1828 & 717 & 13295 & 658 & 86 \\
        Swahili (w/ inst.)   & 5679 & 1828 & 717 & 2247 & 12662 \\\midrule
        Russian (w/o inst.)  & 1828 & 15917 & 6919 & 7056 & 12305 \\
        Russian (w/ inst.)   & 5679 & 15526 & 1828 & 11684 & 7056 \\\midrule
        Thai (w/o inst.)  & 1828 & 717 & 13295 & 6919 & 16152 \\
        Thai (w/ inst.)   & 5679 & 14869 & 12335 & 7056 & 2764 \\\midrule
        Vietnamese (w/o inst.)  & 1828 & 7056 & 6919 & 5549 & 15917 \\
        Vietnamese (w/ inst.)   & 1828 & 5679 & 3579 & 11684 & 717 \\\midrule
        \bottomrule
    \end{tabular}
    }
    \caption{Top-5 activated SAE features per language, with and without instruction}
    \label{tab:language_sae_matrix}
\end{table*}

\begin{table*}[!htb]
    \centering
    \resizebox{0.85\textwidth}{!}{
    \begin{tabular}{c|c}
        \toprule
        \textbf{Language} & \textbf{Unique SAE Features from Instruction ($F^k_\text{inst} \setminus F^k_\text{std}$)} \\
        \midrule
        Arabic & 6356, 5679 \\
        Bengali & 6874, 5679, 31, 6098 \\
        German & 12251, 5679, 15917 \\
        Persian & 16267, 13341, 5679 \\
        Hindi & 7056, 5679, 31, 6098 \\
        Italian & 4427, 5679, 14174 \\
        Swahili & 5679, 2247, 12662 \\
        Russian & 5679, 15526, 11684 \\
        Thai & 5679, 14869, 12335, 2764 \\
        Vietnamese & 5679, 3579, 11684 \\
        \bottomrule
    \end{tabular}
    }
    \caption{SAE features uniquely activated by instruction across languages ($F^{k}_\text{inst} \setminus F^{k}_\text{std}$)}
    \label{tab:sae_instruction_unique}
\end{table*}

\paragraph{Feature-Level Activation Analysis across Languages}

To examine the effect of Instruction following across multilingual prompts, we collect the top-5 SAE features activated for each of the 10 target languages, with and without instruction. As shown in Table~\ref{tab:language_sae_matrix}, several core features—such as $\text{SAE}_{1828}$ (API references), $\text{SAE}_{717}$ (structured text), and $\text{SAE}_{7056}$ (scientific concepts)—appear across nearly all languages and conditions, suggesting shared backbone representations, which are regarded as redundant features and are not used for steering.

Specifically we compute the set difference $F^k_\text{inst}(L) \setminus F^k_\text{std}(L)$ for each language $L$ and present the results in Table~\ref{tab:sae_instruction_unique}. These uniquely activated SAE features reflect dimensions enhanced only under instruction-followed settings. Some, such as $\text{SAE}_{5679}$, appear frequently across languages and may encode general-purpose Instruction following behavior. Others, like SAE $\text{SAE}_{12251}$ (German) or $\text{SAE}_{14869}$ (Thai), may capture more language-specific structural or formatting behaviors.

\subsection{Case V: Instruction with Response Length (Sentence Level)}

In this case, we still include the table, omitting the textual explanations of each useful SAE feature in the length constraint case and instead report the corresponding SAE feature indices using cumulative methods. As mentioned before, we include the instructions in the input when performing steering.

\begin{table*}[!htb]
    \centering
    \resizebox{\textwidth}{!}{
    \begin{tabular}{c|ccccc}
        \toprule
        \midrule
        \textbf{Length} & \textbf{SAE Rank = 1} & \textbf{SAE Rank = 2} & \textbf{SAE Rank = 3} & \textbf{SAE Rank = 4} & \textbf{SAE Rank = 5} \\
        \midrule
        1 (w/o inst.) & 1828 & 717 & 6919 & 15917 & 7056\\
        1 (w/ inst.)  & 1828 & 7056 & 9214 & 2967 & 14983 \\ \midrule
        2 (w/o inst.)  & 1828 & 717 & 6919  &15917 & 7056 \\
        2 (w/ inst.)   & 1828 & 7056 & 9214 & 2967 & 15917\\ \midrule
        3 (w/o inst.)  & 1828 & 717 & 6919  &15917 & 7056 \\
        3 (w/ inst.)   & 1828 & 7056 & 9214 & 15917 & 2967 \\ \midrule

        \bottomrule
    \end{tabular}
    }
    \caption{Top 1-5 activated SAE features for length constraint}
    \label{tab:language_sae_matrix_2}
\end{table*}
\begin{table*}[!htb]
    \centering
    \resizebox{\textwidth}{!}{
    \begin{tabular}{c|ccccc}
        \toprule
        \midrule
        \textbf{Length} & \textbf{SAE Rank = 6} & \textbf{SAE Rank = 7} & \textbf{SAE Rank = 8} & \textbf{SAE Rank = 9} & \textbf{SAE Rank = 10} \\
        \midrule
        1 (w/o inst.) & 5549 & 208 & 12122  & 9214 & 13295 \\
        1 (w/ inst.)  & 15917 & 5679 & 2992 & 3862 & 1377 \\ \midrule
        2 (w/o inst.)  &  5549 & 208 & 12122  & 9214 & 13295\\
        2 (w/ inst.)   & 1377 & 717 & 6919 & 4371 & 2992 \\ \midrule
        3 (w/o inst.)  &  5549 & 208 & 12122  & 9214 & 13295 \\
        3 (w/ inst.)   & 6919 & 1377 & 717 & 4371 & 2992  \\ \midrule

        \bottomrule
    \end{tabular}
    }
    \caption{Top 6-10 activated SAE features for length constraint}
    \label{tab:language_sae_matrix_3}
\end{table*}

\paragraph{Feature-Level Activation Analysis with Response Length Control}

We examine how response length control at the sentence level affects internal SAE activations, comparing top-10 features with and without instruction across target lengths (1–3 sentences). As shown in Tables~\ref{tab:language_sae_matrix_2} and~\ref{tab:language_sae_matrix_3}, the non-instruction setting yields identical feature activations across all lengths, indicating that the base model does not inherently modulate internal activations in response to subtle differences in expected output length.

In contrast, Instruction following introduces new activated SAE features—most notably $\text{SAE}_{9214}$  and S$\text{SAE}_{2967}$ across all lengths. These features are not present in the non-instruction setting and are consistently ranked within the top 5, suggesting they play a central role in sentence-level response control.
$\text{SAE}_{7056}$ , which is shared across both settings, appears to contribute general linguistic structure, while features like $\text{SAE}_{9214}$  and $\text{SAE}_{2967}$ likely encode finer-grained length or formatting control.

\begin{table*}[!htb]
    \centering
    \begin{minipage}{\textwidth}
        \centering
        \resizebox{\textwidth}{!}{%
            \begin{tabular}{lc|c|c|c|c}
                \toprule
                Model Mode & With Instruction  & \multicolumn{1}{c|}{Sentences (Length = 1)} & \multicolumn{1}{c|}{Sentences (Length = 2)} & \multicolumn{1}{c|}{Sentences (Length = 3)} & \multicolumn{1}{c|}{Sentences (Length = 4)} \\ 
                \midrule 
                Inference &$\times$  & 9.30 & 6.98 & 9.30 & 2.33 \\
                MeanActDiff & $\checkmark$  & 74.42 & 67.44 & 60.47 & 51.16\\ 
               $\text{SAE}_{\text{7056}}$ & $\checkmark$ &      69.77 & 58.14 & 58.14 & 55.81\\
               $\text{SAE}_{\text{9214}}$ & $\checkmark$ &      69.77 & 62.79 & 58.14 & 58.14\\
                $\text{SAE}_{\text{2967}}$ & $\checkmark$ &      72.09 & 62.79 & 60.47 & 53.49\\
                $\text{SAE}_{\text{1377}}$ & $\checkmark$ &      74.42 & 62.79 & 65.12 & 51.16\\
                Inference & $\checkmark$ &  72.09 & 60.47 & 58.14 & 51.16\\

                \bottomrule
            \end{tabular}%
        }
        \caption{Gemma-2-9B-it: Performance on Instruction Following Dataset IFEval (From At Least to Exactly)}
        \label{tab:gemma-2-9b_length-exactly}
    \end{minipage}
\end{table*}

\begin{table*}[!htb]
    \centering
    \begin{minipage}{\textwidth}
        \centering
        \resizebox{\textwidth}{!}{%
            \begin{tabular}{lc|c|c|c|c}
                \toprule
                Model Mode & With Instruction  & \multicolumn{1}{c|}{Sentences (Length = 1)} & \multicolumn{1}{c|}{Sentences (Length = 2)} & \multicolumn{1}{c|}{Sentences (Length = 3)} & \multicolumn{1}{c|}{Sentences (Length = 4)} \\ 
                \midrule 
                Inference &$\times$  & 0 & 4.55 & 4.55 & 2.27 \\
                MeanActDiff & $\checkmark$  & 77.28 & 70.46 & 72.73 & 65.91\\ 
               $\text{SAE}_{\text{7056}}$ & $\checkmark$ &  65.91 & 65.91 & 70.46 & 63.64 \\
               $\text{SAE}_{\text{9214}}$ & $\checkmark$ & 72.73 & 68.19 & 63.64 & 65.91\\
                $\text{SAE}_{\text{2967}}$ & $\checkmark$ &  77.28  & 72.73 & 72.73 & 75.00 \\
                $\text{SAE}_{\text{1377}}$ & $\checkmark$ & 70.46 & 63.64 & 63.64   & 70.45  \\
                Inference & $\checkmark$ &  72.73 & 72.73 & 65.91 & 65.91\\
                
                \bottomrule
            \end{tabular}%
        }
        \caption{Gemma-2-9B-it: Performance on Instruction Following Dataset IFEval (From Less Than to Exactly)}
        \label{tab:gemma-2-9b_length-exactly2}
    \end{minipage}
\end{table*}

\section{SAE Feature Similarity}
Since the dimension of SAE steering vectors and MeanActDiff steering vectors are the same, we here define the similarity metrics, with either 1) the internal-similarity between two SAE vectors $\text{SAE}_i$ and $\text{SAE}_j$ :
\begin{equation}
    \text{sim\_I}(\text{SAE}_i, \text{SAE}_j) = \frac{\langle\text{SAE}_i, \text{SAE}_j\rangle}{\| \text{SAE}_i\|_2 \| \text{SAE}_i\|_2}
\end{equation}

or 2) the external-similarity between a pair of $\text{SAE}_i$ and $\text{MeanActDiff}_i$ for each reasoning case:

\begin{equation}
    \text{sim\_E} = \frac{\langle\text{SAE}_i, \text{MeanActDiff}_i\rangle}{\| \text{SAE}_i\|_2 \|\text{MeanActDiff}_i\|_2}
\end{equation}

For clarity, we present two similarity matrices showing the top-5 most activated SAE features for the Gemma-2-2B and Gemma-2-9B models in the math reasoning cases respectively. Furthurmore, we provide a table that reports the similarity metric $\text{sim\_E}$ for Gemma-2-2B and Gemma-2-9B models. This metric quantifies the alignment between the vectors that trigger Chain-of-Thought (CoT) reasoning behavior and those derived from MeanActDiff steering, helping to explain the observed similarities in their effects.

\begin{figure*}[!htb]
    \centering
    \begin{minipage}[!htb]{0.49\linewidth}
        \centering
        \includegraphics[width=\linewidth]{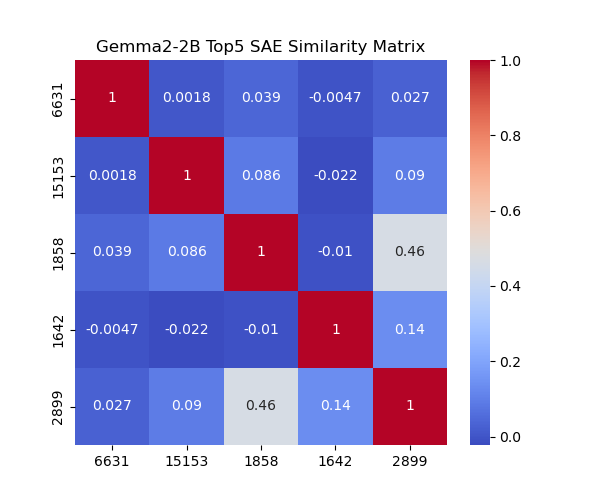}
        \caption*{(a) Gemma-2-2B Top-5 SAE Feature Similariry Matrix}
    \end{minipage}
    \hfill
    \begin{minipage}[!htb]{0.49\linewidth}
        \centering
        \includegraphics[width=\linewidth]{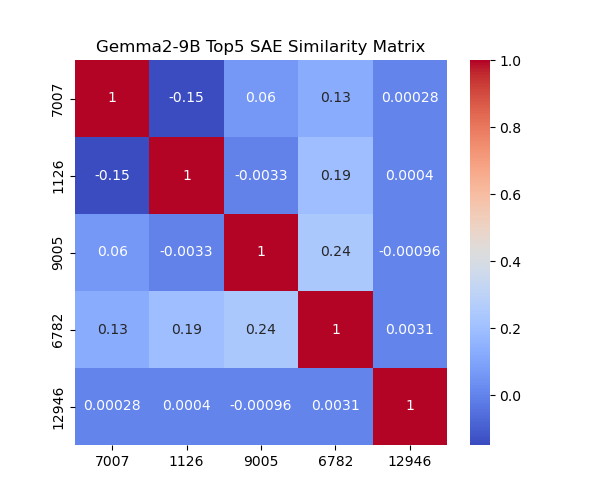}
        \caption*{(b) Gemma-2-9B Top-5 SAE Feature Similariry Matrix}
    \end{minipage}
    \caption{Cosine similarity matrices of Top-5 activated SAE features from Gemma-2-2B and Gemma-2-9B models.}
    \label{fig:gemma-similarity}
\end{figure*}

Overall the similarity between a pair of SAE features is low. This makes sense since SAE training tried to disentangle feature representation. Feature $\text{SAE}_{2899}$ and feature $\text{SAE}_{6782}$ are positively related other four features. $\text{SAE}_{1858}$ and   $\text{SAE}_{2899}$ share high similarity as observed, and in the Neuronpedia we found that they share the common top activated word "The". The similarity between SAE steering vectors and MeanActDiff steering vectors is somehow correlated, which is shown in the Table \ref{tab:cos_sim_cot}.

\begin{table}[!htb]
    \centering
    \begin{tabular}{c|c}
    Model & Sim\_E (max = 1) \\ \midrule
       Gemma-2-2B & 0.2227 \\
       Gemma-2-9B  & 0.3465 \\
    \end{tabular}
    \caption{Cosine similarity between the SAE steering vector that represents CoT reasoning and corresponding MeanActAdd steering vector}
    \label{tab:cos_sim_cot}
\end{table}

\subsection{Instruction Following}

\begin{figure}[!htb]
    \centering
    \includegraphics[width=\linewidth]{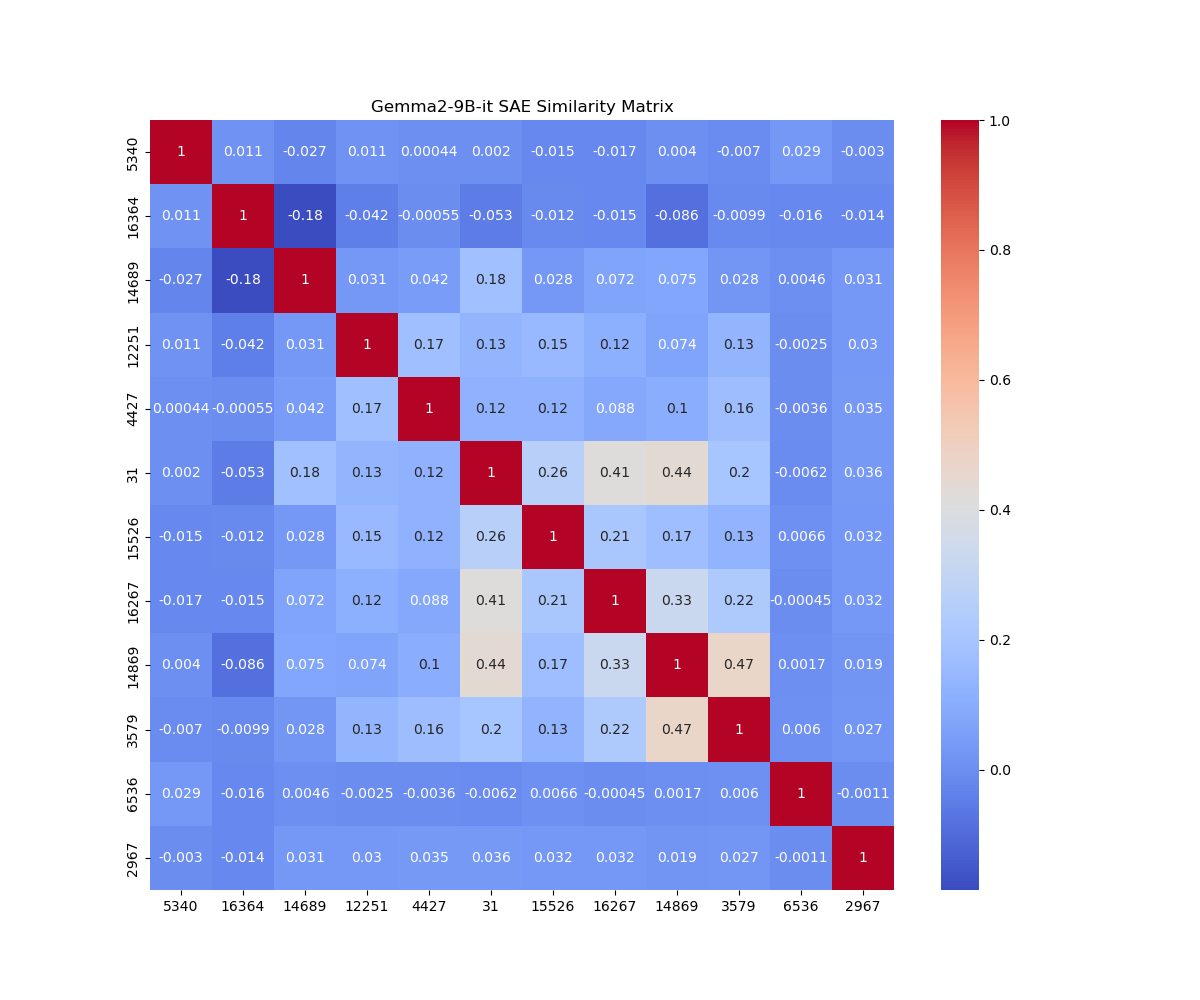}
    \caption{SAE feature similarity matrix}
    \label{fig:9b_it_feature_matrix}
\end{figure}

\begin{figure}[!htb]
    \centering
    \includegraphics[width=\linewidth]{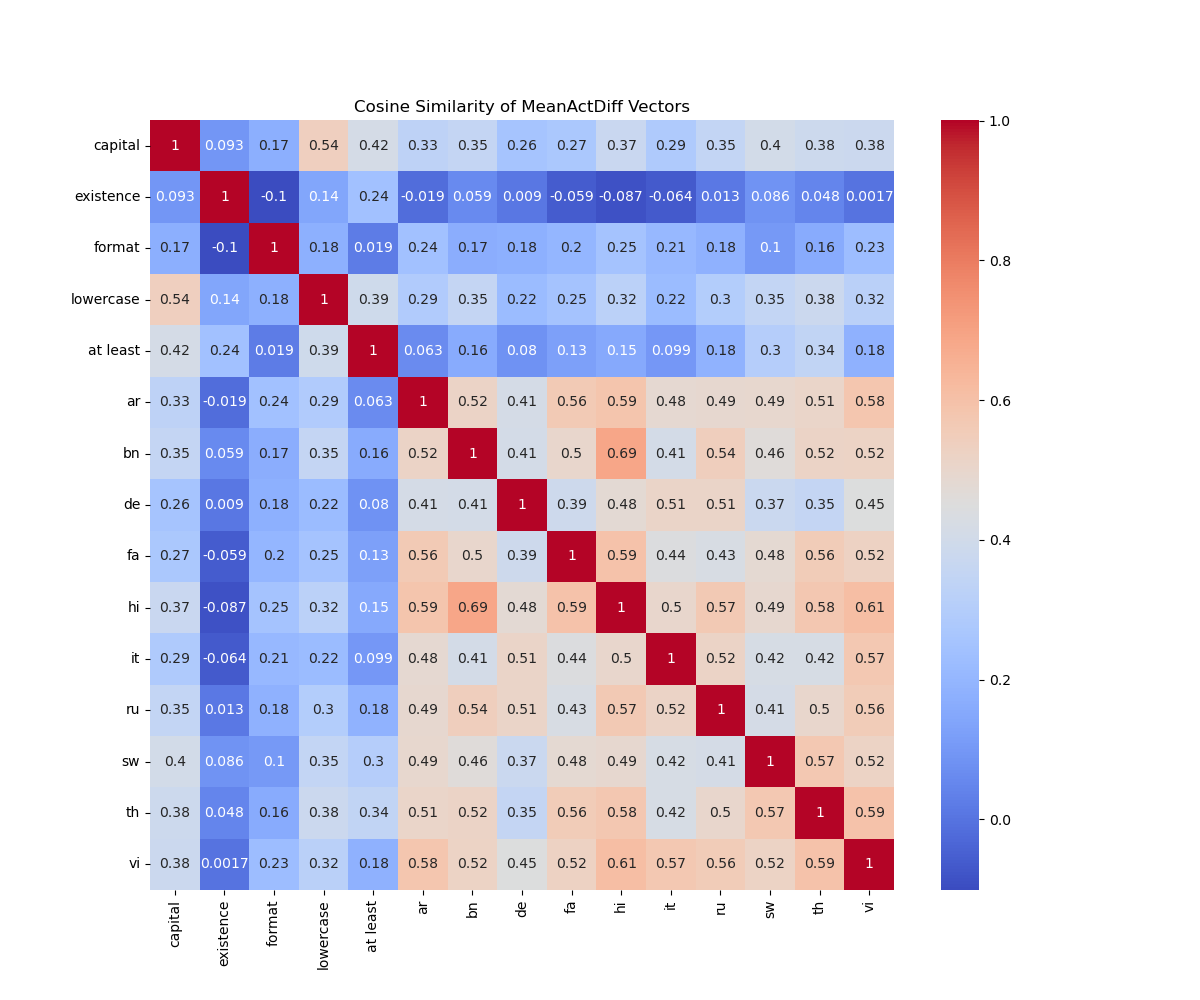}
    \caption{MeanActDiff similarity matrix}
    \label{fig:9b_it_feature_matrix_mean}
\end{figure}

We visualize the pairwise cosine similarity of selected SAE steering vectors (Figure \ref{fig:9b_it_feature_matrix}) and MeanActDiff steering vectors (Figure \ref{fig:9b_it_feature_matrix_mean}), re-using the similarity metrics from section 5.2.3.

In Figure~\ref{fig:9b_it_feature_matrix}, we observe that the cosine similarity between most SAE features is small and close to zero, suggesting that these features are relatively independent. However, a few pairs such as $\text{SAE}_{14869}$ and $\text{SAE}_{3579}$, exhibit moderate similarity (approximately 0.47), indicating some degree of entanglement or functional overlap. $\text{SAE}_{14869}$ corresponds to Thai in the SAE latent space, while $\text{SAE}_{3579}$ represents Vietnamese. The observed similarity may reflect shared characteristics such as geographical proximity, tonal language structure, and linguistic patterns common to Southeast Asian languages, including subject–verb–object syntax and the use of classifiers.

The MeanActDiff vectors show a much clearer semantic structure. Specifically, steering vectors corresponding to \texttt{bn} (Bengali) and \texttt{hi} (Hindi) exhibit high cosine similarity ($>$0.6), which aligns with the linguistic proximity of these languages. This phenomenon suggests that the SAE has learned to group language-related activations in a semantically meaningful way. 

We also observe that the vectors for \texttt{capital} and \texttt{lowercase} have relatively high similarity ($>$0.5), indicating that these features may encode closely related formatting or syntactic styles. This structured similarity does not emerge as clearly in the raw SAE feature space, further confirming that \textbf{MeanActDiff steering surfaces latent semantic relationships more explicitly}, which is the potential reason why MeanActDiff steering is better than SAE steering in lowercase generation.

\begin{table}[!htb]
    \centering
    \begin{tabular}{c|c}
    \toprule
    Case & Sim\_E (max = 1) \\ \midrule
       Uppercase  & 0.4938\\
       Lowercase  & 0.5031\\
       JSON & 0.4460 \\
       Length Constraint & 0.0985\\
       Language-German & 0.4649 \\
       Language-Italian & 0.5209 \\
       Language-Hindi & 0.3030 \\
       Language-Russian & 0.3618 \\
       Language-Persian & 0.4531 \\
       Language-Arabic & 0.4794 \\
       Language-Thai & 0.3470 \\
       Language-Vietnam & 0.3855\\

       \bottomrule
    \end{tabular}
    \caption{Cosine similarity between the SAE steering vector and corresponding MeanActAdd steering vector for each case in instruction-followed IFEval dataset}
    \label{tab:cos_sim}
\end{table}

Table~\ref{tab:cos_sim} reports the cosine similarity (\texttt{Sim\_E}) between the SAE steering vector and its corresponding MeanActAdd steering vector across various instruction cases in the IFEval dataset. A higher similarity indicates that the two steering strategies are capturing consistent directional influence in activation space.

We observe that \textbf{format-related tasks}, such as \textit{Uppercase}, \textit{Lowercase}, and \textit{JSON}, all exhibit relatively high similarity scores (ranging from 0.44 to 0.50), suggesting that these syntactic modifications activate similar directions in the model, regardless of the steering method.
In contrast, \textit{Length Constraint} matching shows the lowest similarity (0.0985), possibly because it involves stricter token-level constraints that are harder to align in latent space, or because the SAE may not localize the exact-matching behavior as strongly.

Among language-related instructions, most language pairs exhibit relatively high cosine similarity (typically above 0.45), indicating that the corresponding MeanActDiff vectors activate overlapping regions in the model’s representation space. For example, vectors corresponding to \textit{Italian}, \textit{Arabic}, and \textit{Persian} show strong alignment, but even languages such as \textit{Hindi}, \textit{Thai}, and \textit{Vietnamese} maintain moderately high similarity with others. This suggests a shared latent structure across language steering signals. The slight variability may be attributed to differences in tokenization coverage, script diversity, or activation sparsity associated with each language.

\section{Prompt Template}

This section outlines the prompt template used in instruction-following tasks. These templates enforce strict formatting and response structures to encourage model consistency.

\vspace{1em}

\begin{table}[!htb]
    \centering
    \caption{Instruction examples for different tasks.}
    \label{tab:instruction_examples}
    \begin{tabularx}{\columnwidth}{@{} l >{\raggedright\arraybackslash}X @{}}
        \toprule
        \textbf{Instruction} & \textbf{Example} \\
        \midrule
        \rowcolors{2}{white}{gray!10} % 从数据行开始交替颜色
        Lowercase & Please ensure that your response is in English, and in all lowercase letters. \\
        Language & Please respond using only the German language, no other language is allowed. \\
        JSON Format & Wrap the entire output in JSON format. \\
        English Capitalize & Make sure your entire response is in English, and in all capital letters. \\
        Sentence Frequency & Answer with exactly 3 sentences. \\
        Word Inclusion & Please include the word cat in the response. \\
        \bottomrule
    \end{tabularx}
\end{table}

For math reasoning prompts, we include 8-shot chain-of-thought examples inside a shaded gray box preceding the query to improve multi-step reasoning:

\vspace{1em}

\begin{tcolorbox}[colback=lightgray, colframe=gray!40, title=Few-shot Chain-of-Thought Examples, breakable]
Question: Olivia has \$23. She bought five bagels for \$3 each. How much money does she have left?  
Answer: Olivia had 23 dollars.\\She bought 5 bagels at 3 dollars each: 5 × 3 = 15.\\She spent 15 dollars, so she has 23 - 15 = 8 dollars left.\\The answer is 8.\\

Question: There were nine computers in the server room. Five more computers were installed each day, from monday to thursday. How many computers are now in the server room? \\
Answer: There were originally 9 computers.\\5 computers were added each day for 4 days: 5 × 4 = 20.\\So, total is 9 + 20 = 29.\\The answer is 29.\\  

Question: There are 15 trees in the grove. Grove workers will plant trees in the grove today. After they are done, there will be 21 trees. How many trees did the grove workers plant today?\\  
Answer: There were originally 15 trees.\\After planting, there were 21 trees.\\So, 21 - 15 = 6 trees were planted.\\The answer is 6.\\  

Question: If there are 3 cars in the parking lot and 2 more cars arrive, how many cars are in the parking lot?  \\
Answer: There were 3 cars.\\2 more cars arrived: 3 + 2 = 5.\\The answer is 5.\\  

Question: Jason had 20 lollipops. He gave Denny some lollipops. Now Jason has 12 lollipops. How many lollipops did Jason give to Denny?  \\
Answer: Jason had 20 lollipops.\\After giving some away, he had 12.\\So he gave away 20 - 12 = 8.\\The answer is 8.\\  

Question: Shawn has five toys. For Christmas, he got two toys each from his mom and dad. How many toys does he have now?  \\
Answer: Shawn had 5 toys.\\He got 2 from mom and 2 from dad: 2 + 2 = 4.\\So he has 5 + 4 = 9 toys.\\The answer is 9.\\  

Question: Leah had 32 chocolates and her sister had 42. If they ate 35, how many pieces do they have left in total?  \\
Answer: Leah had 32 and her sister had 42: 32 + 42 = 74.\\They ate 35: 74 - 35 = 39.\\The answer is 39.\\

Question: Michael had 58 golf balls. On tuesday, he lost 23 golf balls. On wednesday, he lost 2 more. How many golf balls did he have at the end of wednesday?  \\
Answer: Michael started with 58.\\Lost 23 on Tuesday: 58 - 23 = 35.\\Lost 2 more on Wednesday: 35 - 2 = 33.\\The answer is 33.\\

\end{tcolorbox}

\section{Simple Evaluation/Verification with Visualization}
\textbf{Neuronpedia}~\citep{neuronpedia} is an interactive web-based platform that hosts labeled SAE feature dictionaries for open-weight models. Each SAE feature is tagged with human-interpretable descriptions (e.g., ``list start'', ``closing brace'', ``polite language'') based on automated annotation pipelines by gpt 4o. We utilize Neuronpedia to investigate and verify the semantics of key SAE latents identified during steering experiments. Specifically, we validate whether activation-based interventions correspond to meaningful internal concepts.

\begin{figure}[!htb]
    \centering
    \includegraphics[width=\linewidth]{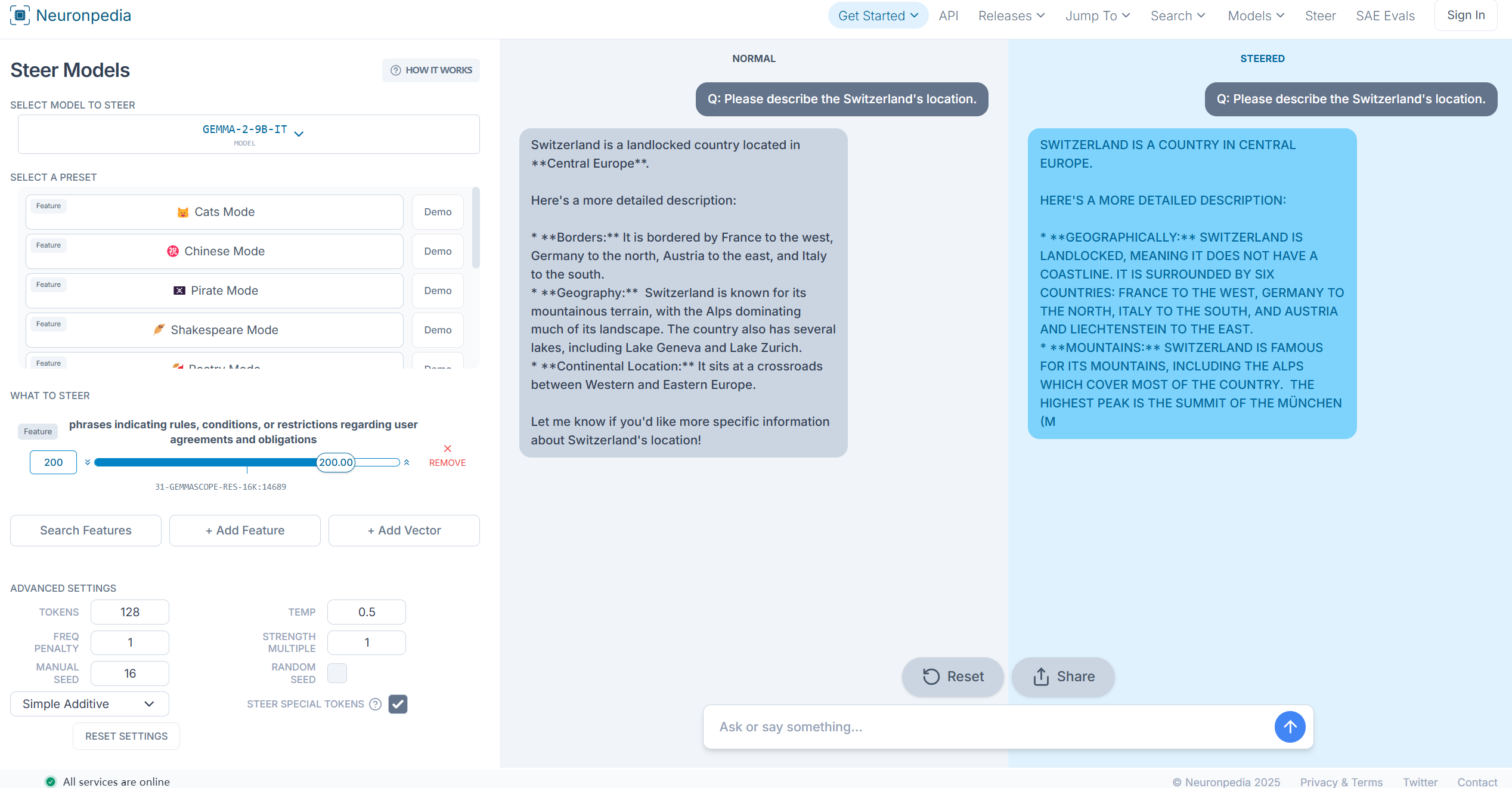}
    \caption{Example 1 by Neuronpedia: Steering models to generate uppercase answers by SAE steering vectors}
    \label{fig:steering_sae_1}
\end{figure}

\begin{figure}[!htb]
    \centering
    \includegraphics[width=\linewidth]{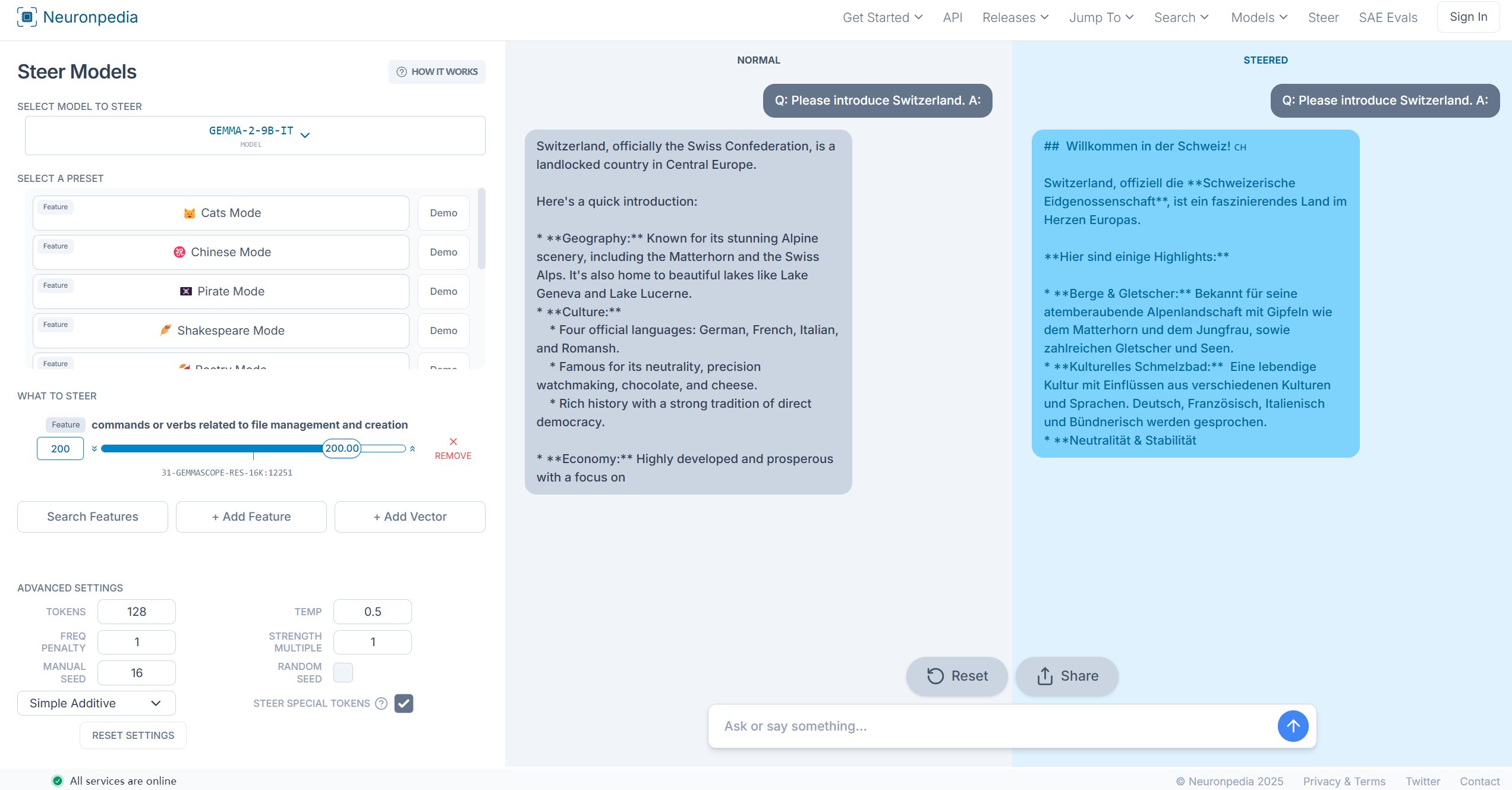}
    \caption{Example 2 by Neuronpedia: Steering models to generate German answers by SAE steering vectors}
    \label{fig:steering_sae_2}
\end{figure}

\section{Linear Probing with SAE Feature Gating}

\begin{figure*}
    \centering
    \includegraphics[width=\linewidth]{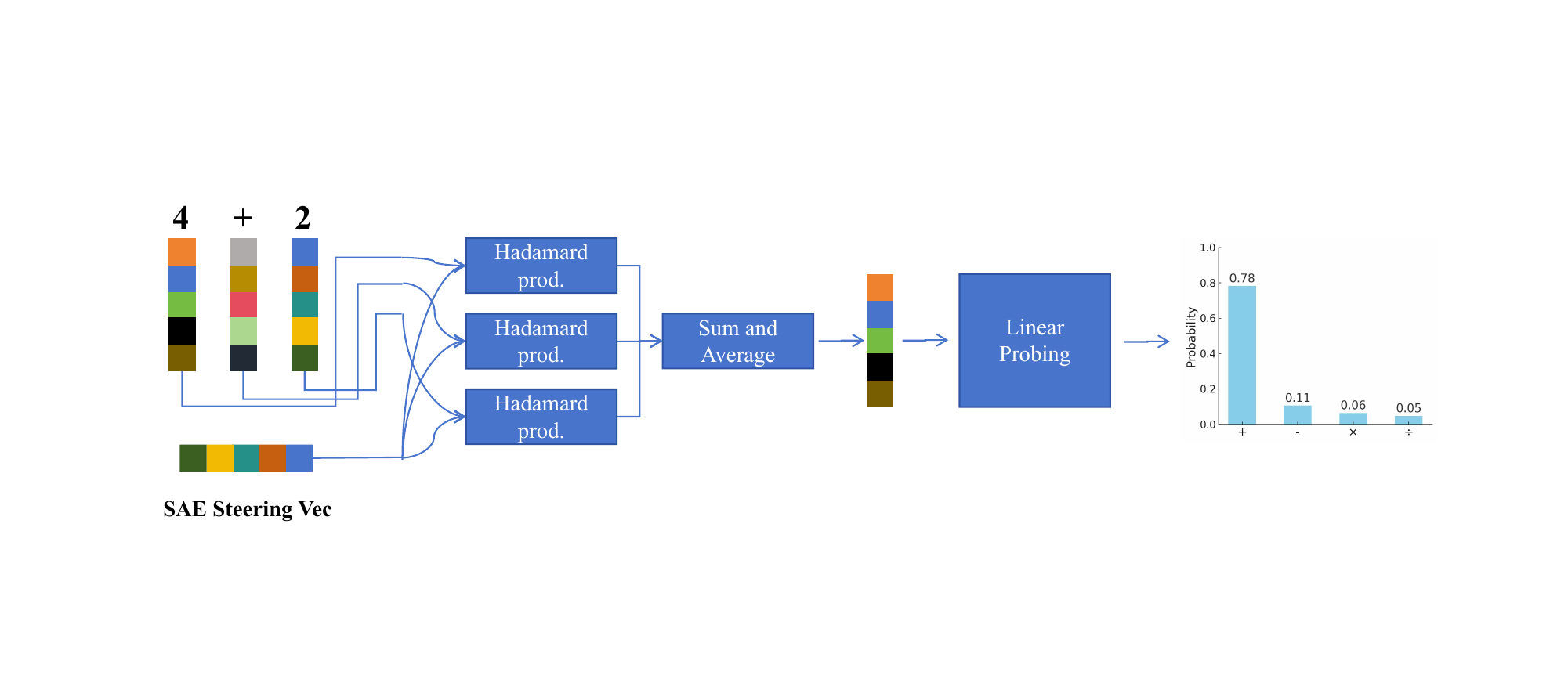}
    \caption{Linear Probing}
    \label{fig:linear_probing_methods}
\end{figure*}

\subsection{Methods}

To quantitatively assess whether specific SAE features encode concepts related to arithmetic, we employ a linear probing methodology. The objective is to train a simple linear classifier to decode a specific arithmetic operation (addition, subtraction, multiplication, or division) from the internal representations of a language model. The performance of this probe serves as a measure of how linearly separable the target information is within the model's activations.

Our probing setup introduces a key modification to the standard approach by using an SAE feature vector as a gating mechanism. For a given input sequence representing an arithmetic expression, we first extract the sequence of hidden state activations, $\mathbf{H}$, from a target layer of the frozen language model. Instead of probing these activations directly, we modulate them via an element-wise Hadamard product ($\odot$) with a chosen SAE feature vector, $\mathbf{d}_{l,j}$:
\begin{equation}
    \mathbf{H}_{\text{gated}} = \mathbf{H} \odot \mathbf{d}_{l,j}
\end{equation}
This operation selectively amplifies or suppresses dimensions in the activation space that are aligned with the feature captured by the SAE vector.

The resulting sequence of gated activation vectors, $\mathbf{H}_{\text{gated}}$, is then aggregated into a single vector representation through mean-pooling across the token dimension. This final vector is fed into a linear classifier, which is trained to output logits for the four operation classes. A softmax function is applied to these logits to produce a probability distribution. The classification accuracy of this probe indicates the extent to which the specific SAE feature $\mathbf{d}_{l,j}$ encodes information relevant to the arithmetic operation being classified.

\subsection{Results}

\begin{minipage}[t]{0.48\textwidth}
\captionsetup{hypcap=false}
\captionof{table}{Probing with SAE 15153 and Gemma-2-2B}
\begin{tabular}{lccc}\toprule
 & precision & recall & f1-score \\\midrule
+ & 0.33  &   0.41   &   0.36 \\
- & 0.23  &   0.29   &   0.26 \\
* & 0.55  &   0.21   &   0.31 \\
/ & \bf 0.72  &  \bf 0.82   &   \bf0.76\\
\midrule
macro avg & 0.46   &   0.43  &    0.42 \\ \bottomrule
\end{tabular}
\end{minipage}

\begin{minipage}[t]{0.48\textwidth}
\captionsetup{hypcap=false}
\captionof{table}{Probing with SAE 1642 and Gemma-2-2B}
\begin{tabular}{lccc}\toprule
 & precision & recall & f1-score \\\midrule
+ & \bf 0.44   & \bf  0.56    & \bf 0.49 \\
- & 0.29   &   0.19    &  0.23 \\
* & 0.38   &   0.36    &  0.37 \\
/ & 0.42   &   0.47    &  0.44 \\
\midrule
macro avg &  0.38   &   0.40   &   0.38\\ \bottomrule
\end{tabular}
\end{minipage}

\vspace{1cm}

\noindent
\begin{minipage}[t]{0.48\textwidth}
\captionsetup{hypcap=false}
\captionof{table}{Probing with SAE 435 and Gemma-2-2B}
\begin{tabular}{lccc}\toprule
 & precision & recall & f1-score \\\midrule
+ & 0.77   &  \bf 0.87   &  \bf 0.82 \\ 
- & 0.55    &  0.43   &   0.48 \\
* & 0.50   &   0.52    &  0.51\\
/ & \bf0.78   &   0.83    &  0.81 \\
\midrule
macro avg & \bf0.65   &  \bf 0.66   &  \bf 0.65\\ \bottomrule
\end{tabular}
\end{minipage}

\begin{minipage}[t]{0.48\textwidth}
\captionsetup{hypcap=false}
\captionof{table}{Probing with MeanActDiff Steering Vector and Gemma-2-2B}
\begin{tabular}{lccc} \toprule
 & precision & recall & f1-score \\\midrule
+ &0.30    &  0.11   &   0.16 \\
- & 0.43   &   0.89    &  0.58 \\
* & 0.72   &   0.14    &  0.23 \\
/ & \bf0.67   & \bf  0.94    & \bf 0.78 \\
\midrule
macro avg & 0.53    &  0.52   &   0.44\\ \bottomrule
\end{tabular}
\end{minipage}

\vspace{1em}
\noindent
\begin{minipage}[t]{0.48\textwidth}
\captionsetup{hypcap=false}
\captionof{table}{Probing with SAE 7007 and Gemma-2-9B}
\begin{tabular}{lccc} \toprule
 & precision & recall & f1-score \\ \midrule
+ & 0.41  &    0.76   &   0.53 \\
- & 0.11  &    0.05   &   0.07 \\
* & 0.78  &    0.33   &   0.47 \\
/ &\bf 0.71  & \bf  0.94    &  \bf 0.81\\
\midrule
macro avg &\bf 0.50    &  \bf0.52  &  \bf  0.47 \\ \bottomrule
\end{tabular}
\end{minipage}

\begin{minipage}[t]{0.48\textwidth}
\captionsetup{hypcap=false}
\captionof{table}{Probing with SAE 9005 and Gemma-2-9B}
\begin{tabular}{lccc} \toprule
 & precision & recall & f1-score \\ \midrule
+ & 0.34   &   0.75    &  0.47 \\
- & 0.42   &   0.15    &  0.22 \\
* & 0.46   &   0.04    &  0.07 \\
/ & \bf0.67   &  \bf 0.91    &\bf  0.77 \\
\midrule
macro avg & 0.47 &     0.46   &   0.38\\ \bottomrule
\end{tabular}
\end{minipage}

\vspace{1cm}

\noindent
\begin{minipage}[t]{0.48\textwidth}
\captionsetup{hypcap=false}
\captionof{table}{Probing with SAE 6782 and Gemma-2-9B}
\begin{tabular}{lccc} \toprule
 & precision & recall & f1-score \\ \midrule
+ & 0.45  &    0.75  &    0.56\\ 
- &  0.46  &    0.25  &    0.33 \\
* & 0.36   &   0.17   &   0.23 \\
/ &\bf 0.67   &  \bf 0.89   &  \bf 0.77 \\
\midrule
macro avg & 0.49   & \bf  0.52    & \bf 0.47  \\ \bottomrule
\end{tabular}
\end{minipage}

\begin{minipage}[t]{0.48\textwidth}
\captionsetup{hypcap=false}
\captionof{table}{Probing with MeanActDiff Steering Vector and Gemma-2-9B}
\begin{tabular}{lccc}\toprule
 & precision & recall & f1-score \\\midrule
+ & 0.45   &   0.75    &  0.56 \\
- & 0.24   &   0.15    &  0.18 \\
* & 0.72   &   0.22    &  0.34  \\
/ &\bf 0.57   &  \bf 0.80    &\bf  0.67 \\
\midrule
macro avg & 0.49   &   0.48   &   0.44\\ \bottomrule
\end{tabular}
\end{minipage}

\vspace{2em}
As shown in Tables 27 through 32, the linear probing results reveal notable differences in performance across various SAE steering vectors. These experiments are conducted on synthetic data specifically designed around basic arithmetic operations, including addition, subtraction, multiplication, and division. As a result, SAE vectors that encode operation-specific information such as $\text{SAE}_{435}$ and $\text{SAE}_{7007}$) tend to achieve higher probing accuracy, with macro-average F1-scores of 0.65 and 0.47, respectively.

In contrast, SAE vectors like $\text{SAE}_{15153}$, previously shown to be effective in generating complex Chain-of-Thought reasoning or instruction-following behaviors, achieve lower probing scores with macro F1-score of 0.42. This suggests that their activation patterns are better aligned with higher-level reasoning rather than surface-level operation classification.  This suggests that SAE features specialized for basic symbolic manipulation are more effective for linear probing on arithmetic-focused datasets, whereas reasoning-related SAE vectors are less suited for such tasks due to their reasoning nature.

\end{document}